\documentclass[10pt,twocolumn,letterpaper]{article}

\usepackage{cvpr}              
\usepackage[accsupp]{axessibility}

\usepackage{graphicx}
\usepackage{amsmath}
\usepackage{amssymb}
\usepackage{booktabs}
\usepackage{xcolor}
\usepackage{xfrac}
\usepackage{multirow}
\usepackage{soul}
\usepackage{adjustbox}
\usepackage{pgf}
\usepackage{wrapfig}

\newcommand{\myparagraph}[1]{\textbf{#1}}

\newcommand{\pci}{$\mathcal{S}_{t}$\xspace}
\newcommand{\pcj}{$\mathcal{S}_{t+1}$\xspace}

\newcommand{\floxels}{Floxels\xspace}

\definecolor{cvprblue}{rgb}{0.21,0.49,0.74}
\usepackage[pagebackref,breaklinks,colorlinks,allcolors=cvprblue]{hyperref}

\def\paperID{15297}
\def\confName{CVPR}
\def\confYear{2025}

\title{Floxels: Fast Unsupervised Voxel Based Scene Flow Estimation}

\author{David T. Hoffmann$^{*,1,2}$ \quad
Syed Haseeb Raza$^{*,3}$ \quad
Hanqiu Jiang$^{*,1}$ \quad \and
Denis Tananaev$^1$ \quad
Steffen Klingenhoefer$^{3}$ \quad
Martin Meinke$^{*,1}$\\
$^*$Equal contribution \quad
$^1$Robert Bosch GmbH\quad
$^2$University of Freiburg \quad $^3$CARIAD SE\\
{\tt\small syed.haseeb.raza@cariad.technology\quad hanqiu.jiang@de.bosch.com}
}

\begin{document}

\maketitle

\begin{abstract}
Scene flow estimation is a foundational task for many robotic applications, including robust dynamic object detection, automatic labeling, and sensor synchronization. Two types of approaches to the problem have evolved: 1) Supervised and 2) optimization-based methods. Supervised methods are fast during inference and achieve high-quality results, however, they are limited by the need for large amounts of labeled training data and are susceptible to domain gaps. In contrast, unsupervised test-time optimization methods do not face the problem of domain gaps but usually suffer from substantial runtime, exhibit artifacts, or fail to converge to the right solution. In this work, we mitigate several limitations of existing optimization-based methods. To this end, we 1) introduce a simple voxel grid-based model that improves over the standard MLP-based formulation in multiple dimensions and 2) introduce a new multi-frame loss formulation. 3) We combine both contributions in our new method, termed Floxels. On the Argoverse 2 benchmark, Floxels is surpassed only by EulerFlow among unsupervised methods while achieving comparable performance at a fraction of the computational cost. Floxels achieves a massive speedup of more than $\sim60-140\times$ over EulerFlow, reducing the runtime from a day to 10 minutes per sequence. 
Over the faster but low-quality baseline, NSFP, Floxels achieves a speedup of $\sim 14 \times$.
\end{abstract}
\section{Introduction}
\label{sec:intro}

\begin{figure*}[t]
    \centering

    \parbox[b]{0.02\textwidth}{\rotatebox{90}{~~~~~~~\textbf{FNSF \cite{li2023fast}}}}  
    \hspace{0.005\textwidth}
    \begin{subfigure}[b]{.22\textwidth}
        \centering
        \includegraphics[width=1\textwidth,trim={1cm 4cm 3cm 6cm}, clip]{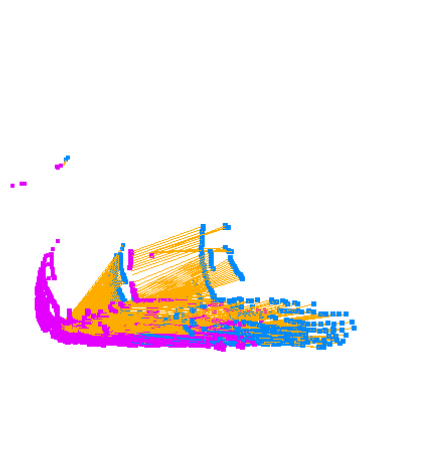}
    \end{subfigure}
    \hspace{0.013\textwidth}
    \begin{subfigure}[b]{.22\textwidth}
        \centering
        \includegraphics[width=1\textwidth,trim={0.75cm 0.5cm 0.75cm 0.5cm}, clip]{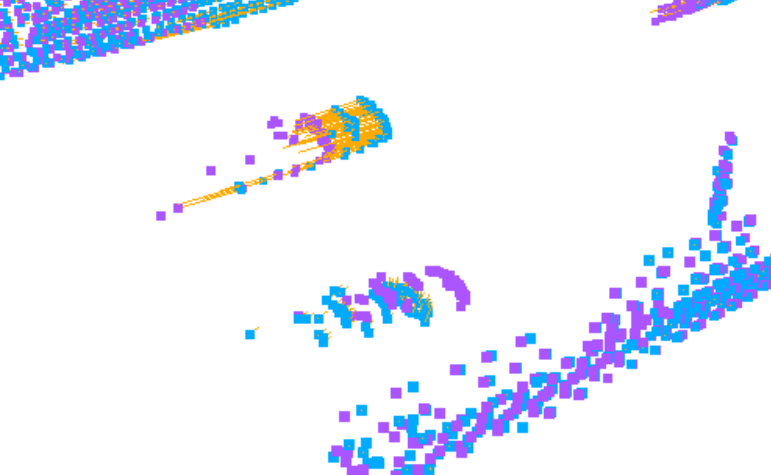}
    \end{subfigure}
    \hspace{0.013\textwidth}
    \begin{subfigure}[b]{.22\textwidth}
        \centering
        \includegraphics[width=1\textwidth,trim={3cm 0cm 0 0cm}, clip]{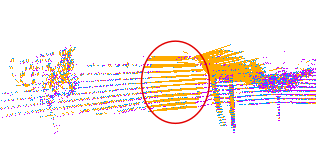}
    \end{subfigure}
    \hspace{0.013\textwidth}
    \begin{subfigure}[b]{.22\textwidth}
        \centering
        \includegraphics[width=1\textwidth,trim={1.0cm 1.0cm 3cm 1.0cm}, clip]{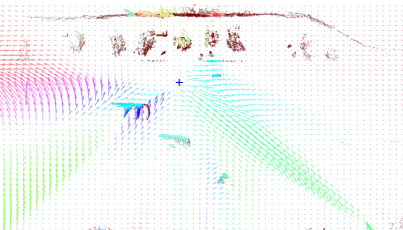}
    \end{subfigure}    

    \vspace{0.25cm}

    \parbox[b]{0.02\textwidth}{\rotatebox{90}{~~~~~~~~~~~\textbf{Floxels (ours)}}}  
    \hspace{0.005\textwidth}
    \begin{subfigure}[b]{.22\textwidth}
        \centering
        \includegraphics[width=1\textwidth,trim={1cm 3cm 3.5cm 6cm}, clip]{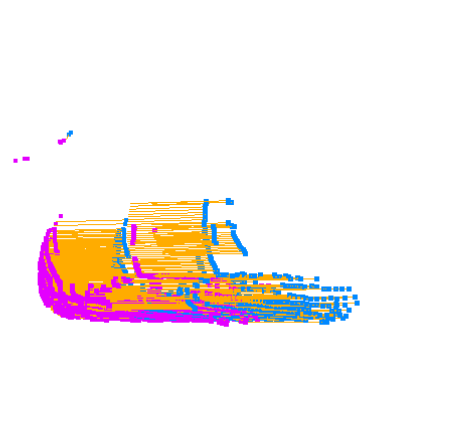}
        \caption{Estimated flow for a car. FNSF is biased towards close points.}
        \label{fig:teaser_near_points}
    \end{subfigure}
    \hspace{0.013\textwidth}
    \begin{subfigure}[b]{.22\textwidth}
        \centering
        \includegraphics[width=1\textwidth,trim={0.75cm 0.5cm 0.75cm 0.5cm}, clip]{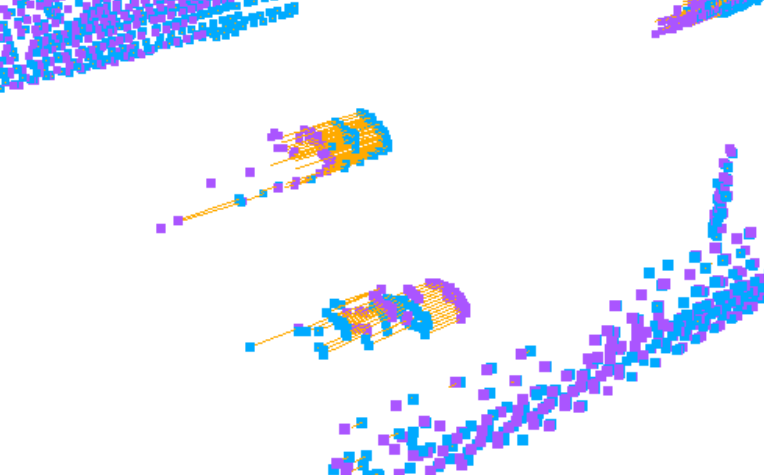}        
        \caption{Cars moving in opposite directions. FNSF misses some motion.}
        \label{fig:teaser_opp_dir}
    \end{subfigure}    
    \hspace{0.013\textwidth}
    \begin{subfigure}[b]{.22\textwidth}
        \centering
        \includegraphics[width=1\textwidth,trim={3cm 0cm 0 0cm}, clip]{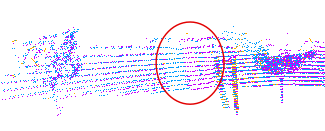}          
        \caption{Static Wall with occlusion. Floxels correctly predicts no flow.}
        \label{fig:teaser_occlusion}
    \end{subfigure}
    \hspace{0.013\textwidth}
    \begin{subfigure}[b]{.22\textwidth}
        \centering
        \includegraphics[width=1\textwidth,trim={1.0cm 1.0cm 3cm 1.0cm}, clip]{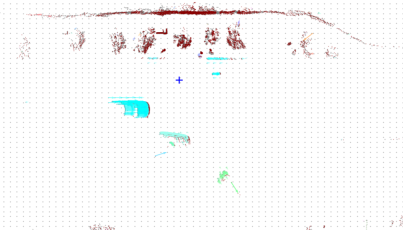}
        \caption{Birds-eye flow field with false flow in empty regions for FNSF.}
        \label{fig:teaser_flow_field}
    \end{subfigure}        
    \caption{\textbf{Examples of scene flow} from \textbf{Fast Neural Scene Flow (FNSF) \cite{li2023fast} (top)} and \textbf{our method \floxels (bottom)}. We show {\color{RubineRed}point cloud $t_1$ (Purple)}, {\color{RoyalBlue}point cloud $t_2$ (blue)} and the {\color{BurntOrange}{estimated scene flow (orange)}} in Fig.~\ref{fig:teaser_near_points}, \ref{fig:teaser_opp_dir}, \ref{fig:teaser_occlusion}. 
    Fig.~\ref{fig:teaser_near_points}: FNSF tends to predict flow to closest points. \floxels uses neighboring points to escape such local minima.
    Fig.~\ref{fig:teaser_occlusion}: A static wall occluded by trees. FNSF predicts flow in the {\color{BrickRed}occluded region (red circles)} to the nearest points. \floxels overcomes this by making use of multiple scans.
    Fig.~\ref{fig:teaser_flow_field}: Birds-eye view on the flow field. FNSF displays wrong flow patterns in regions without objects. \floxels correctly predicts zero flow for these regions.}
    \label{fig:teaser}
\end{figure*}

We live in a dynamic, three-dimensional world. To operate safely in this complex environment, any autonomous agent must perceive its three-dimensional structure and dynamics. This is reflected in many tasks, such as moving object detection, trajectory prediction, camera-lidar synchronization, collision avoidance, or point cloud densification.
Sensors that provide information about the structure of the 3D world are widespread nowadays, particularly in the form of Lidar scanners found in cars, robots, phones, and augmented reality devices. However, estimating the dynamics of a scene remains a challenge. 
A common formulation of the problem is referred to as \emph{scene flow estimation}: 
The task of estimating a dense 3D motion field that represents how points move in space across consecutive lidar scans.

As diverse as the devices and applications that require scene flow estimation are, so are the proposed algorithms. Existing scene flow algorithms can be coarsely sorted into two categories, both of which come with an individual set of advantages and disadvantages. \emph{(Semi-)supervised methods} \cite{liu2019flownet3d, wu2020pointpwcnetcoarsetofinenetworksupervised, SLIM, puy2020flotsceneflowpoint, pontes2020sceneflowpointclouds,zhang2024deflowdecodersceneflow,khatri2025can,kim2024flow4dleveraging4dvoxel} and \emph{optimization-based approaches} \cite{li2021neural, li2023fast, vedder2024zeroflowscalablesceneflow, lin2024icpflowlidarsceneflow, zhang2024seflowselfsupervisedsceneflow, vedder2024neuraleuleriansceneflow}.
(Semi-)Supervised methods tend to be fast and perform well. However, they require large and expensive datasets for training and large GPUs even for inference. Additionally, they demonstrate the typical issues associated with learning-based methods when exposed to domain shifts, such as new environments or variations in motion characteristics.
In this work, we will focus on test-time optimization methods. These methods are limited mainly by their significantly longer inference time \cite{li2021neural, vedder2024neuraleuleriansceneflow}, but they promise high-quality results, generalization capabilities by design and can be used to generate pseudo ground truth data for supervised methods \cite{vedder2024zeroflowscalablesceneflow}.
However, as we will show, current methods often fall short of delivering the anticipated levels of quality (see Fig.~\ref{fig:teaser}) or require massive amounts of compute \cite{vedder2024neuraleuleriansceneflow}. 

In this work, we will address both problems of previous methods. 
While EulerFlow \cite{vedder2024neuraleuleriansceneflow} exhibits stunning results, the enormous computational demand limits its practical use. 
The alternatives are lightweight optimization-based methods like NSFP, but they usually lead to low-quality flow.
To overcome these limitations we first analyze the methods NSFP \cite{li2021neural} and FNSF \cite{li2023fast}. 
Our analysis reveals that MLP-based methods exhibit a ``homogeneous motion" bias in the shadow of objects (Fig.~\ref{fig:teaser_occlusion}) and predict wrong flow patterns in empty regions, referred to as ``windmill artifacts'', as they resemble the sails of a windmill (see Fig.~\ref{fig:teaser_flow_field}).
Last, we observe false flow associations, when a dynamic point is closer to a false point than to its true counterpart (Fig.~\ref{fig:teaser_near_points}).
While using EulerFlow might resolve such problems, the enormous computational demand will prohibit its use in most scenarios.
But do we really need to rely on slowly converging time-conditioned MLPs to obtain high-quality scene flow? Do we have to optimize over hundreds of point clouds to obtain good results? 
We find that both are unnecessary for achieving competitive results.
We observe that a time-conditioned NSFP is not required, demonstrate that a few point clouds are sufficient, and propose a simpler method that significantly accelerates runtime while maintaining high-quality results.

In particular, to overcome the limitations of NSFP and EulerFlow, we propose a new method called Scene \textbf{Flo}w V\textbf{oxels} (\floxels), which mitigates the issues mentioned. To enhance convergence characteristics over NSFP, improve overall results, and fix the windmill artifacts, we replace the MLP with a simple voxel grid.
To overcome the problem caused by missing corresponding points in adjacent point clouds, we extend the problem to more than a single scan.
Even if the corresponding points are missing in some scans, they are likely present in adjacent scans.
To avoid cross-object point matching, we employ a clustering-based loss, which encourages points close in space to move in similar directions. We choose the parameters to potentially yield several clusters per object to prevent false cluster assignment.
\floxels makes use of multiple timesteps using Euler integration but does not require the training of a time-conditioned representation. This leads to more than $60-140\times$ decrease of runtime over EulerFlow while resulting in competitive flow estimates.
Thus, \floxels improves over classical test-time optimization methods both in accuracy and runtime and improves over the recent method EulerFlow \citep{vedder2024neuraleuleriansceneflow} drastically in runtime while performing competitively. \floxels even performs close to the recently proposed high-performing supervised methods, DifFlow3D \cite{liu2024difflow3drobustuncertaintyawarescene} and Flow4D \citep{kim2024flow4dleveraging4dvoxel}.

In summary, our contributions are: 
1) We analyze the failure cases of MLP-based test-time optimization methods like NSFP and FNSF.
2) We demonstrate that these methods exhibit biases, which manifest as windmill artifacts, incorrect flow in the shadow of objects, and a tendency to predict (often incorrect) flow toward the nearest point.
3) To overcome these limitations we introduce multiple novel loss components and extend the optimization to include multiple time steps.
4) To improve convergence speed over NSFP and FNSF and resolve the observed artifacts we replace the MLP with a simple parameterized voxel grid.
5) Among unsupervised methods, our resulting method \floxels is only outperformed by EulerFlow on the Argoverse 2 CVPR 2024 Scene Flow Challenge \citep{khatri2025can} and performs competitively with, or even surpasses, supervised methods.
6) Floxels reduces the gap to the concurrent work EulerFlow, but estimation is more than 60-140$\times$ faster, outpacing even FNSF, with runtime gains over FNSF increasing with point cloud size.

\section{Related works}

\myparagraph{Feedforward methods.}
Historically, scene flow estimation has been treated as a learning problem. The parameters of a model are optimized over training batches of subsequent point clouds to produce the best possible generalization to point cloud sequences observed at inference time. Many popular models are trained in a fully-supervised way \cite{liu2019flownet3d, wu2020pointpwcnetcoarsetofinenetworksupervised,zhang2024deflowdecodersceneflow,khatri2025can}.
Some methods propose to use per-point representations \cite{zhang2024deflowdecodersceneflow} or to simply use an object tracking approach \cite{khatri2025can}.
Another line of research trains in the semi-supervised setting \cite{SLIM, puy2020flotsceneflowpoint, pontes2020sceneflowpointclouds,zhang2024seflowselfsupervisedsceneflow,lin2024icpflowlidarsceneflow} by introducing nearest-neighbor losses and extending them by designing explicit loss functions for dynamic and static points \cite{zhang2024seflowselfsupervisedsceneflow}.
Others combine the two by generating pseudo-ground truth for supervised methods using optimization-based methods \cite{vedder2024zeroflowscalablesceneflow}.

\myparagraph{Implicit representations.}
In contrast to these learning-based methods, implicit representations and coordinate networks learn scene-specific representations via test-time optimization.
Pioneered by NeRF \cite{mildenhall2020nerfrepresentingscenesneural}, which trains an MLP to represent a 3D RGB scene. Runtime improvements were obtained by replacing the MLP with a feature grid \cite{fridovich2022plenoxels,fridovichkeil2023kplanesexplicitradiancefields}.

\myparagraph{Neural scene flow prior.}
Inspired by the works on coordinate-based networks \citet{li2021neural} propose the seminal method NSFP to model 3D scene flow using a coordinate MLP. It compels with the promise that due to the test-time optimization, it generalizes well to novel scenes with different statistics compared to learning-based methods.
However, optimization is expensive, which lead to the follow-up work \textit{Fast Neural Scene Flow} (FNSF) \cite{li2023fast}, where speedups are achieved by replacing a nearest-neighbor-based cost formulation with trilinear interpolation in distance transforms (DT).
In this work, we make use of the speed-ups achieved by FNSF.
However, we identify that unfavorable convergence characteristics of the MLP still result in a bad trade-off between runtime and performance.

Inspired by \cite{fridovich2022plenoxels}, we retrace the path taken by NeRFs and replace the MLP in NSFP with a simple voxel grid.
\citet{li2023fast} also test a more straightforward linear model, which employs a grid to learn flow but they complicate the model by using complex positional embeddings \cite{zheng2022trading}, which require an encoder and a blending function to summarize points. While they do observe speedups, they report lower performance than for FNSF.

\myparagraph{Pair vs.~multi-scan.}
Several recent approaches tackle the problem by optimizing scene flow over longer sequences 
\cite{9879660, liu2024selfsupervisedmultiframeneuralscene, vedder2024neuraleuleriansceneflow}.
This makes the approaches more robust to occlusions and misassociations.
In particular, EulerFlow \citep{vedder2024neuraleuleriansceneflow} enforces flow to be consistent over multiple subsequent scans by learning a time-conditioned NSFP and propagating points through a window of $\pm k$ adjacent scans via Euler integration.
EulerFlow shines when many frames are available, however, the combination of repeated Euler integration, expensive loss computation for a huge number of points, and the slow convergence of the time-conditioned MLP results in enormous computational demand. Using shorter sequence lengths (50 or fewer scans) leads to mediocre performance \cite{vedder2024neuraleuleriansceneflow}.
Our method performs well even with shorter sequence lengths. In comparison, it is incredibly fast and achieves better results than EulerFlow with up to 50 scans, all while requiring fewer frames.

\myparagraph{Loss formulation.}
\label{rel:loss}
Self-supervised and optimization approaches usually use some form of geometric nearest neighbor computation \cite{SLIM, li2021neural, lang2023scoopselfsupervisedcorrespondenceoptimizationbased} as primary loss. 
This comes at a significant cost in runtime since nearest neighbors have to be recomputed after each optimizer step. We adopt the strategy of \cite{li2023fast} and instead constrain the loss using a set of Distance Transforms - one for each scan.
Since most agents in relevant environments are dominated by rigid motion, several types of local rigidity constraints have been proposed. Graph Laplacian regularization on k-NN neighborhoods \cite{pontes2020sceneflowpointclouds} or radius-constrained subgraphs \cite{lang2023scoopselfsupervisedcorrespondenceoptimizationbased} have been been proposed.
Alternative formulations make use of clustering (usually leveraging DBSCAN) to introduce flow constraints within each cluster \cite{najibi2022motion, vidanapathirana2023multi}.
The work by \citet{chodosh2023reevaluatinglidarsceneflow} argues that it is advantageous to introduce the rigidity assumption in a sampling-based postprocessing step. 
In this work, we opt for a cluster-consistency loss.

\section{Methods --- \floxels}

\begin{figure}
    \begin{subfigure}[b]{.52\columnwidth}
        \includegraphics[width=1\textwidth]{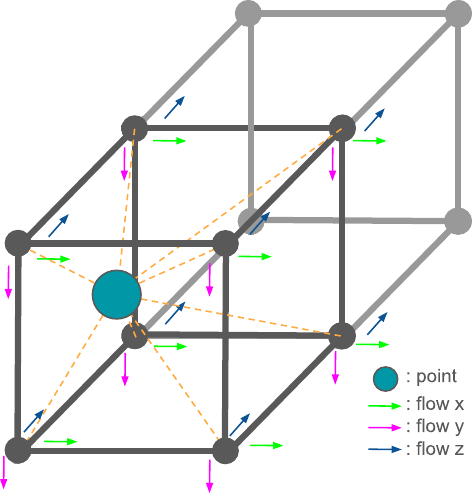}
        \caption{Voxel grid.}
        \label{fig:meth_voxel_grid}
    \end{subfigure}
    \begin{subfigure}[b]{.42\columnwidth}
        \includegraphics[width=1\textwidth]{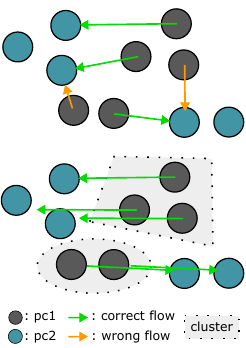}
        \caption{Cluster consistency loss.}
        \label{fig:rigid_body}
    \end{subfigure}
    \begin{subfigure}[b]{.5\textwidth}
        \includegraphics[width=1\columnwidth,trim={0 1.7cm 0 0.75cm},clip]{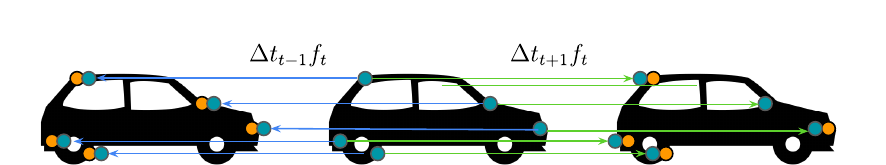}
        \caption{Multi-scan DT loss. We optimize $f_t$, s.t.~errors after projections to different time steps are minimized. We show only 3 time steps but more can be used.}
        \label{fig:meth_constant_velo}
    \end{subfigure}
    \caption{\textbf{Floxel components. }\ref{fig:meth_voxel_grid} \textbf{Voxel Grid.} Instead of an MLP, we use a simple grid to represent the motion of the points. In each voxel (here depicted as vertex), we learn the x,y, and z velocities. For each {\color{TealBlue} point (blue)}, the flow is calculated via trilinear interpolation from the neighboring vertices (connected via {\color{Dandelion}yellow lines}). The final motion is predicted using trilinear interpolation. \ref{fig:rigid_body} \textbf{Cluster consistency loss.} We encourage points of the same cluster to have a similar flow. \ref{fig:meth_constant_velo} \textbf{Multi-scan Distance Transform loss.} To estimate the motion of points at time $t$ we not only rely on the points at $t_{1}$, but also other close-by time points. Thus, even if matching points are missing (car mirror at $t+1$) the flow can be estimated correctly using $t-1$ or more general $t \pm m$.}
    \label{fig:method}
\end{figure}
In this work, we investigate NSFP \citep{li2021neural} and FNSF \citep{li2023fast} in more detail. NSFP \cite{li2021neural} and FNSF \cite{li2023fast} both use an implicit representation implemented by an MLP to represent the scene flow field. They claim that simple MLPs show beneficial regularization properties for scene flow estimation and discourage the use of graph Laplacian-based priors.
We observe that the MLP converges very slowly and identify issues with its ability to regularize flow while at the same time capturing higher-frequency details. Further, we observe that existing solutions suffer when corresponding points are unobservable in one of the two consecutive point clouds or when dynamic objects are near other objects.
In this work, we propose mitigations to all these problems. We summarize them in Fig.~\ref{fig:method}.

Inspired by \cite{fridovich2022plenoxels}, we address the runtime problem by replacing the MLP with a \textbf{voxel grid}, which we find to converge faster and lead to better results.
To mitigate the problems caused by occlusion changes, we constrain flow to be consistent across small sequences of scenes instead of scan pairs. 
To fix the problems caused by close-by points, we add a clustering consistency loss, encouraging clusters of points to have similar motions.
Last, we observe that noisy flow estimates in static regions can be overcome by using a simple l2 regularizer on the flow.
We describe all these components in detail in the following subsections.

\subsection{Voxel grid instead of MLP}

Training an MLP to represent 3D space as an implicit function takes a long time to converge \cite{fridovich2022plenoxels}.
Following \citet{fridovich2022plenoxels}, we replace the MLP with an explicit 3D voxel grid representation. 
Our approach focuses on the problem of scene flow rather than the typical neural rendering problem that NeRFs try to solve, so we parameterize the grid directly with 3D flow vectors, as depicted in Fig.~\ref{fig:meth_voxel_grid}.

\floxels represents the scene as a 3D grid, where each grid corner is assigned a set of parameters directly representing the 3D flow $f_t \in \mathbb{R}^3$. These grid corners can be considered trainable support points to a vector field spanned by trilinear interpolation within each grid cell. This leads to a smooth flow field spanning the entire grid.
In contrast to MLPs' black box characteristics, this representation allows for straightforward interpretation and predictable learning behavior. In particular, gradients propagated for a single point only affect neighboring grid cells.
It also provides a natural regularizer enforcing local smoothness.
The disadvantage of using a grid instead of an MLP is that a large voxel size limits the spatial resolution of the flow field. At the same time, a too-small voxel size leads to longer runtime, increased memory usage, and little regularization.

We find that a voxel size of 0.5 meters yields a good trade-off between speed and accuracy in road scenes.
We train the voxel grid with a learning rate of 0.05 without weight decay for a maximum of 500 epochs and stop optimization using early stopping with a patience of 250 epochs and a minimum delta of $0.01$.

\subsection{\floxels loss}
\label{sec:losses}
\textbf{Basic loss.}
The basic scene flow loss can be described as
\begin{equation}
    \ell_\text{d}= D(\mathcal{S}_t + f_t, \mathcal{S}_{t+1}),
\end{equation}
where \pci and \pcj are two consecutive point clouds, where $t$ and $t+1$ indicate the time point, $f_t$ is the scene flow at time point $t$ and $D(\cdot, \cdot)$ is a distance function assuming that the optimal flow  $f_t$ transforms \pci into \pcj. 
Note that, in the unsupervised setting, we neither have point correspondences between \pci and \pcj nor can we expect \pci and \pcj to contain an equal number of points.
We follow \citet{li2023fast} and use distance transforms (DT) to avoid recomputing explicit nearest neighbors after each optimizer iteration. Instead, we compute the distance for the predicted point cloud $\mathcal{S}_t + f_t$ 
on a 3D distance transform built for $\mathcal{S}_{t+1}$.

\myparagraph{Multi-scan distance transform loss.}
When computing nearest-neighbor-based losses, like the Chamfer distance or DT on pairs of point clouds, changes in occlusion or large translations often lead to wrong associations.
As a remedy to both, instead of using only one ``target'' point cloud, we use additional neighboring point clouds to further constrain the flow estimate, as depicted for one additional point cloud in Fig.~\ref{fig:meth_constant_velo}. 
To this end, we assume constant velocity and
add loss terms for additional supervision from adjacent scans.
More formally, our loss is given by
\begin{equation}
    \ell_\text{d}= \sum_{\substack{-m \leq t \leq m \\ t \neq 0}}^{} \frac{\lambda(t)}{N} D(\mathcal{S}_0 + f_0 \Delta_t, S_t),
\end{equation}
where $\mathcal{S}_0$ is the target point cloud, $f_0$ is the corresponding flow estimate and $\Delta_t$ indicates the time difference between $t$ and the target point cloud. $N$ is the number of points in $\mathcal{S}_t$ and $\lambda(t)$ a time difference dependent weight.
While $D$ can be any distance, here we opt for distance transform (DT) \cite{li2023fast}, i.e.~$D$ describes the mean distance of all points in the propagated point cloud to the DT for $t$. This formulation requires precomputing $2*(m-1)$ distinct distance transforms, one for each support frame.
In practice, we set $\lambda(t)=\sfrac{1}{t^2}$ to decay the contribution of the point clouds with increasing time difference from the reference cloud. This helps to initially steer the optimization to the correct solution. 
In practice, we reject distance transform values beyond 5~meters (for a sampling rate of 10\,Hz) to enhance robustness concerning outliers or changes in occlusion.
Unless otherwise stated, we use five scans.

\myparagraph{Cluster consistency loss.}
When dynamic objects are close to static objects, false associations and, as a result, false flow estimation can happen, as depicted in Fig.~\ref{fig:rigid_body}.
To mitigate this, we follow \cite{najibi2022motion, vidanapathirana2023multi} and cluster the points in $\mathcal{S}_t$ using DBSCAN \citep{ester1996density} to encourage similar flow within each cluster. 
This effectively works against errors caused by false associations and many-to-one associations.
Note that we aim for over-clustering of objects, as it does not lead to catastrophic failure, but joining two objects with different motions into the same cluster can (imagine a car passing by a static scene element at proximity).
To this end, we choose the parameters to over-segment the space, s.t.~each object may consist of multiple clusters. 
We find the DBSCAN parameters $\epsilon=0.5$ and $\text{min\_points}=4$ to yield good results.
Finally, we compute the cluster loss as 
\begin{equation}
    \ell_C= \frac{1}{N}\sum_{i=0}^N ||f_{t}^{i} - f_{C_{i}}||_2,
\end{equation}
where $f_{t}^{i}$ indicates the flow of point $i$ at time step $t$, and $f_{C_{i}}$ is the mean flow of the cluster $C_i$ for which $i \in C_i$.

\myparagraph{Flow regularizer.}
Minor occlusion changes and sampling effects may lead to spurious flow predictions even in static regions. 
To prevent this, we add a minor penalty to the magnitude of all flow values.
\begin{equation}
    \gamma = ||f_t||_2.
\end{equation}

\myparagraph{Final loss.}
Our final loss is given by
\begin{equation}
    \ell = \lambda_{\text{d}}\ell_\text{d} + (2m -1)  (\lambda_{C} \ell_C + \lambda_\gamma \gamma).
\end{equation}
Note that $\ell_\text{d}$ scales with the number of point clouds. We use $(2m -1)$ to scale the other components similarly.

\section{Experiments}

\begin{figure*}[h!]
    \centering
    \begin{adjustbox}{width=1\linewidth, trim=0.025cm 0.0cm 0.75cm 0.8cm, clip}
    \input{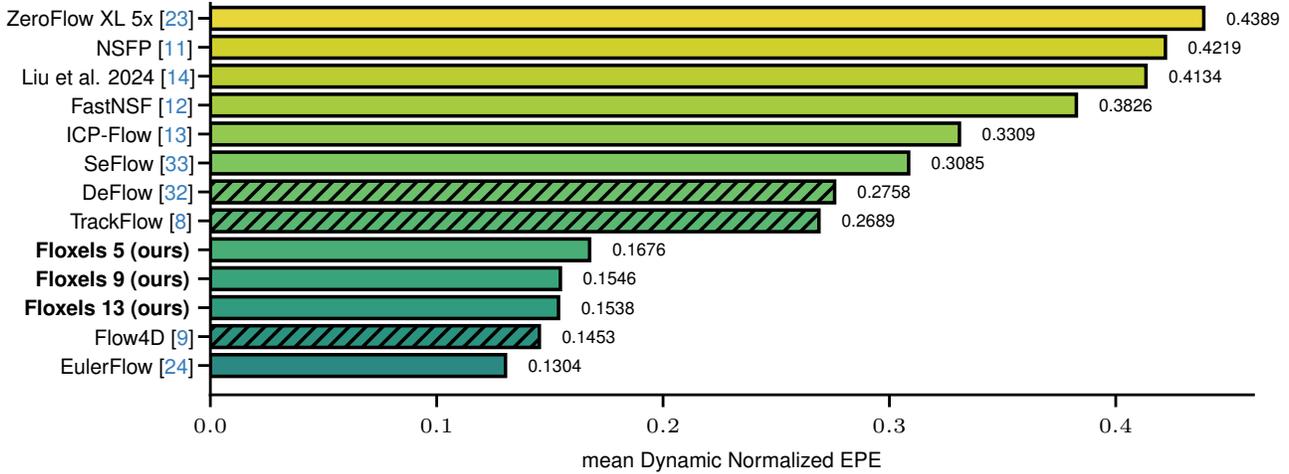}   
    \end{adjustbox}
    \caption{\textbf{Argoverse 2 (2024) Scene Flow Challenge test set.} Mean Dynamic Normalized EPE of Floxels compared to prior art. We report \floxels results for sequence lengths 5, 9, and 13. Supervised methods are shown with hatching. Floxels performs almost as well as EulerFlow, despite requiring only a fraction of computational resources. We show these results also in \cref{tab:bucket_val_full}.}
    \label{fig:mean_average_bucketed_epe}
\end{figure*}

\begin{figure*}
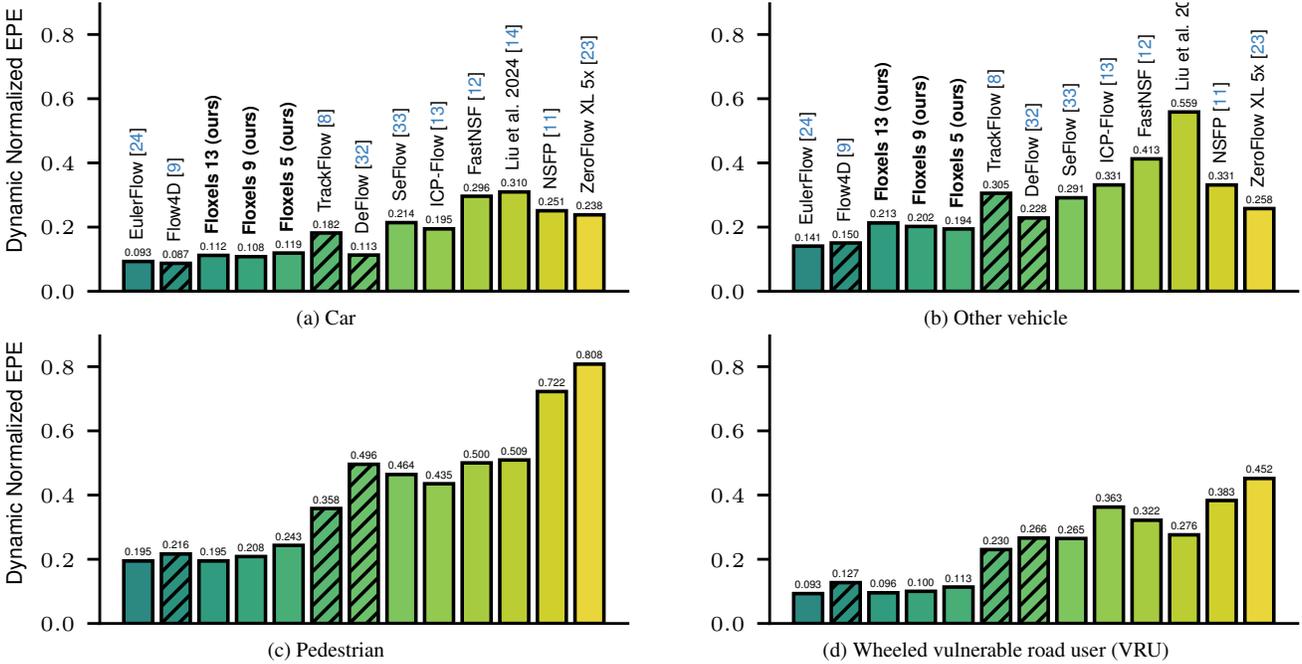

    \centering
    
    \captionsetup[subfigure]{skip=0pt}  

    \begin{subfigure}{0.49\textwidth}
        \centering
        \begin{adjustbox}{width=1\linewidth, trim=0.1cm 0.5cm 0.5cm  2\baselineskip, clip}
        \input{figures/per_metacatagory_bar_CAR_dynamic.pgf}
        \end{adjustbox}
        \caption{Car}
    \end{subfigure}
    \hfill
    \begin{subfigure}{0.49\textwidth}
        \centering
        \begin{adjustbox}{width=1\linewidth, trim=0.1cm 0.5cm 0.5cm  2\baselineskip, clip}
        \input{figures/per_metacatagory_bar_OTHER_VEHICLES_dynamic.pgf}
        \end{adjustbox}
        \caption{Other vehicle}
    \end{subfigure}

    \begin{subfigure}{0.49\textwidth}
        \centering
        \begin{adjustbox}{width=1\linewidth, trim=0.1cm 0.5cm 0.5cm 2\baselineskip, clip}
        \input{figures/per_metacatagory_bar_PEDESTRIAN_dynamic.pgf}
        \end{adjustbox}
        \caption{Pedestrian}
    \end{subfigure}
    \hfill
    \begin{subfigure}{0.49\textwidth}
        \centering
        \begin{adjustbox}{width=1\linewidth, trim=0.1cm 0.5cm 0.5cm 2\baselineskip, clip}
        \input{figures/per_metacatagory_bar_WHEELED_VRU_dynamic.pgf}
        \end{adjustbox}
        \caption{Wheeled vulnerable road user (VRU)}
    \end{subfigure}

    \caption{\textbf{Per-class Dynamic Normalized EPE on Argoverse 2 (2024) Scene Flow Challenge test set.} Supervised methods are shown with hatching. Bars are ordered from left to right by increasing mean Dynamic Normalized EPE.}
    \label{fig:mean_average_bucketed_epe_by_class}
\end{figure*}

\myparagraph{Datasets.} 
We report on Argoverse 2 \cite{wilson2023argoverse2generationdatasets} dataset and follow the protocol proposed by \cite{khatri2025can}. To compare Floxes with NSFP and FNSF in detail and to run ablations we also show results on nuScenes-mini, nuScenes validation \cite{caesar2020nuscenes} and Argoverse validation \cite{argoverse}.
For these datasets, pseudo ground truth scene flow generation mostly follows \cite{li2021neural, li2023fast} and we use a protocol inspired by \citep{chodosh2023reevaluatinglidarsceneflow}.
Thus, we differentiate between the flow errors of static and dynamic points. We categorize points as dynamic if the motion implied by their corresponding object annotations exceeds 0.05~m between subsequent frames.
We provide details in \cref{sec:data_generation_details}.
For \floxels, we optimize the hyperparameters \textit{learning rate, grid-cell size, number of scans, the weights $\lambda_d$, $\lambda_C$, $\lambda_\gamma$} using grid search on nuScenes mini. We do not optimize these hyperparameters for nuScenes validation Argoverse 1 or Argoverse 2, unless otherwise stated. 

\myparagraph{Metrics.}
On Argoverse 2 we report the metrics as proposed by \citet{khatri2025can}, namely, mean dynamic normalized EPE (mdnEPE) as the main metric, which evaluates the percentage of the described motion and tackles the heavy weight of large objects on the standard EPE metrics by averaging over object classes.
We complement this metric by the class-specific dynamic normalized EPE (dnEPE).
For Argoverse 1 and nuScenes we report the same metrics as \cite{li2021neural,li2023fast,liu2019flownet3d,mittal2020just,pontes2020scene,wu2020pointpwc}: 3D \textbf{end-point error (EPE)}, which measures the mean euclidean distance between predicted and ground truth flow, \textbf{$\text{Acc}_{5}$}, which accepts a transformed point, if the EPE $<$ 0.05m or EPE' $<$ 5\%. Here, EPE' is the relative error.
\textbf{$\text{Acc}_{10}$} is defined accordingly with larger threshold.
Last, we report the angle error, i.e., the angle between predicted and ground truth vectors.

\myparagraph{Baseline details.}
To compare on Argoverse 2 we report numbers as provided on the official leaderboard.
On nuScenes and Argoverse 1 we evaluate 3 baselines to compare to Floxels. Among the optimization-based baselines, we chose NSFP \cite{li2021neural} and FNSF \cite{li2023fast}.
As a supervised baseline, we compare to the high-performing recently proposed method DifFlow3D \cite{liu2024difflow3drobustuncertaintyawarescene}. 
We test DifFlow3Ds generalization, i.e., without fine-tuning.

\subsection{Main results}

\begin{table}[t]
    \caption{\textbf{Dynamic points on Argoverse 1 test set}. Models without early stopping (5000 epochs) denoted with ``*''. ``-N'' indicates the number of layers. Results for static points are shown in \cref{tab:argo_full}.}
    \label{tab:argo}
    \centering
    \resizebox{\columnwidth}{!}{
    \begin{tabular}{l|c c c c | c}
    \toprule
    Method & EPE $\downarrow$ & $\text{Acc}_{5}$ $\uparrow$ & $\text{Acc}_{10}$ $\uparrow$ & angle error $\downarrow$ & Time (s)\\
         \midrule
         & \multicolumn{4}{c|}{Test-time optimization (identical early stopping)} \\
         \midrule
         NSFP-8 & 0.200 & 0.288 & 0.521 & 0.468  & 63.01 \\
         NSFP-16 & 0.226 & 0.280 & 0.498 & 0.530 &  72.54 \\
         FNSF-8 & 0.282 & 0.281 & 0.518 & 0.588 & 20.66 \\         
         \floxels (5s) & \textbf{0.104} & \textbf{0.537} & \textbf{0.755} & \textbf{0.420} & \textbf{4.38} \\
         \midrule         
         & \multicolumn{4}{c|}{Test-time optimization} \\
         \midrule
         NSFP-8* & 0.202 & 0.272 & 0.508 & 0.478 &  -- \\ 
         NSFP-16* & 0.203 & 0.336 & 0.541 & 0.495 &  -- \\
         FNSF-8* & 0.370 & 0.215 & 0.458 & 0.651 & -- \\ 
         \floxels (5s) & \textbf{0.109} & \textbf{0.526} & \textbf{0.739} & \textbf{0.423} & -- \\ 
         \bottomrule
    \end{tabular}
    }
\end{table}
Our main results are summarized in Figures \ref{fig:mean_average_bucketed_epe} and \ref{fig:mean_average_bucketed_epe_by_class}.
As \cref{fig:mean_average_bucketed_epe} reveals Floxels performs second best among the unsupervised methods, only outperformed by EulerFlow, however, EulerFlow is more than $\sim60-140\times$ slower than Floxels.
Floxels even outperforms recent supervised methods \cite{zhang2024deflowdecodersceneflow,khatri2025can} and performs almost on par with the best supervised method Flow4D \cite{kim2024flow4dleveraging4dvoxel}. As \cref{fig:mean_average_bucketed_epe_by_class} shows, Floxels outpreforms Flow4D for \textit{Pedestrians} and \textit{Wheeled VRU}. For pedestrians, Floxels even performs on par with EulerFlow.

On nuScenes and Argoverse 1 we observed the best tradeoff between accuracy and runtime for 5 scans, and thus report numbers for 5 scans, however, as shown in \cref{tab:abl_nr_frames} more scans improve results. 
Among the self-supervised methods, \floxels performs best across all metrics on dynamic points (\cref{tab:nusc_test_full,tab:argo}) 
and better or on par on static points (\cref{tab:nusc_test_full,tab:argo_full}).
Improvements over previous methods are particularly pronounced for the dynamic points across all metrics and datasets. Note that the hyperparameters are only tuned for nuScenes-mini but generalize well.

NSFP and FNSF perform poorly on dynamic points. On Argoverse 1 and nuScenes we tested two factors that could explain the poor results. First, the training might stop prematurely. We train all baselines without early stopping to factor this out, which does not help.
Second, a larger network might improve results. Thus, we also run NSFP with 16 layers, which leads to no or mild improvements.

The supervised method DifFlow3D \cite{liu2024difflow3drobustuncertaintyawarescene} generalizes to nuScenes and is a viable option on small-scale point clouds as used in nuScenes. However, the memory requirement of DifFlow3D scales with the number of points, s.t.~already for Argoverse 1 it requires more than 16~GB of GPU memory and exceeds our available GPU memory.
\begin{figure}[t]
    \centering
    \begin{subfigure}[b]{1.0\columnwidth}
        \centering
        \includegraphics[width=0.33\textwidth]{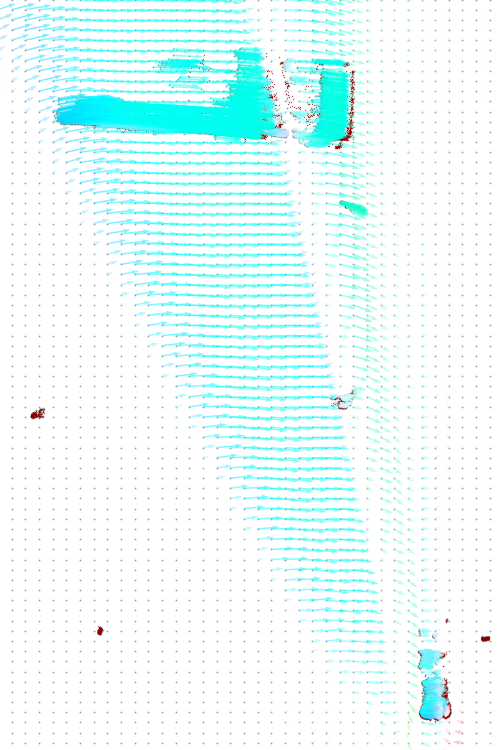}
        \includegraphics[width=0.66\textwidth]{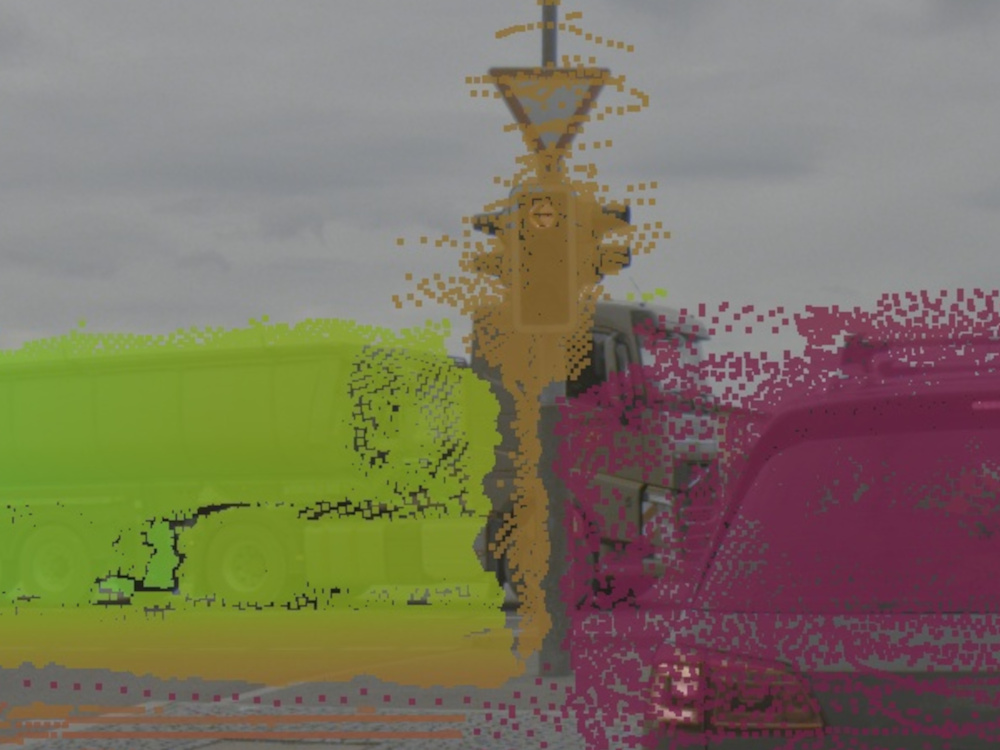}
        \caption{MLP}
    \end{subfigure}    
    \begin{subfigure}[b]{1.0\columnwidth}
        \centering
        \includegraphics[width=0.33\textwidth]{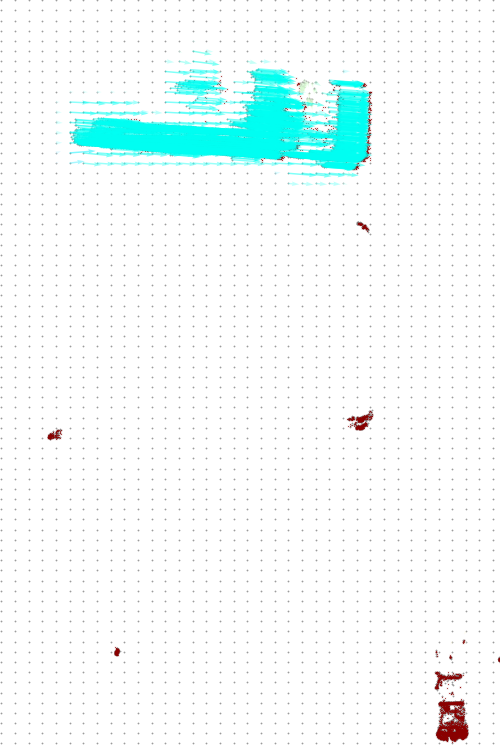}
        \includegraphics[width=0.66\textwidth]{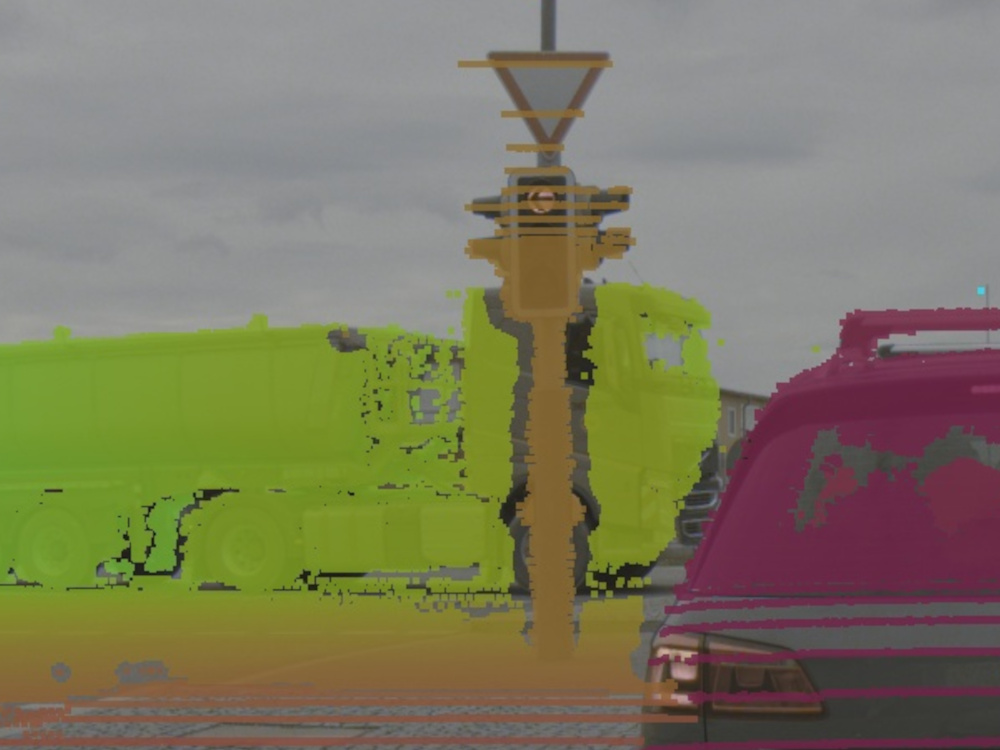}
        \caption{Floxels}
    \end{subfigure}
    \caption{\textbf{Left: Birds-eye view of the flow field.}
    A Truck passing behind a traffic light. The neural prior leads to a prediction of false flow in empty regions (windmill artifacts), and no flow is predicted for occluded regions. Windmill artifacts do not contribute to the loss metrics in regions without actual points, resulting in an overestimated performance for MLP-based methods. \textbf{Right: Accumulated point clouds projected to the camera.} Windmill artifacts of the MLP lead to points of the static car being falsely shifted to the left during lidar-to-camera synchronization. \floxels are not susceptible to this failure mode.}
    \label{fig:birds_eye_flow}
\end{figure}

\begin{figure}[t]
    \centering
    \begin{subfigure}[b]{1.0\columnwidth}
        \centering
        \includegraphics[width=0.32\textwidth,trim={0.39cm 0 0.25cm 0.25cm},clip]{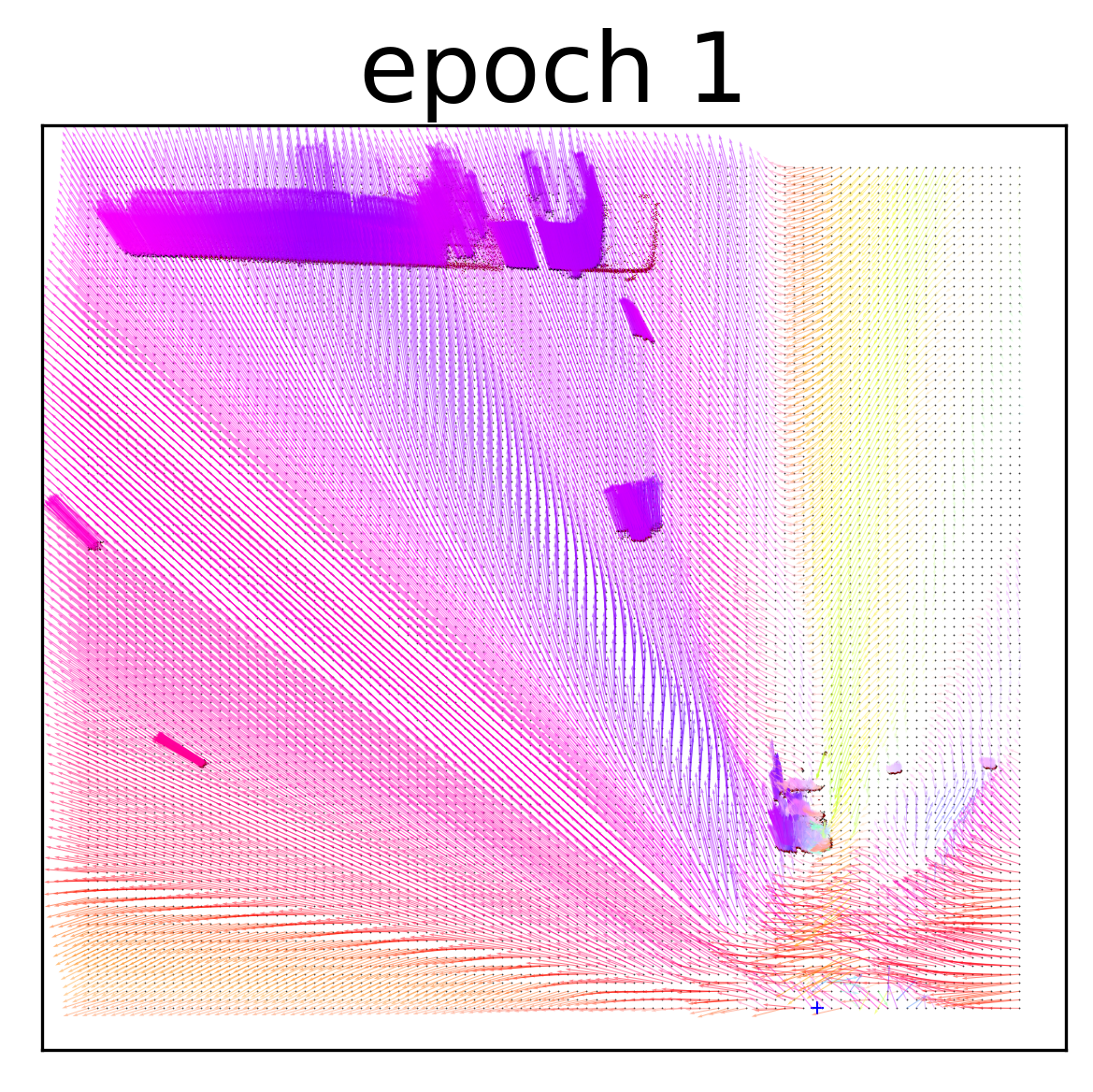}
        \includegraphics[width=0.32\textwidth,trim={0.39cm 0 0.25cm 0.25cm},clip]{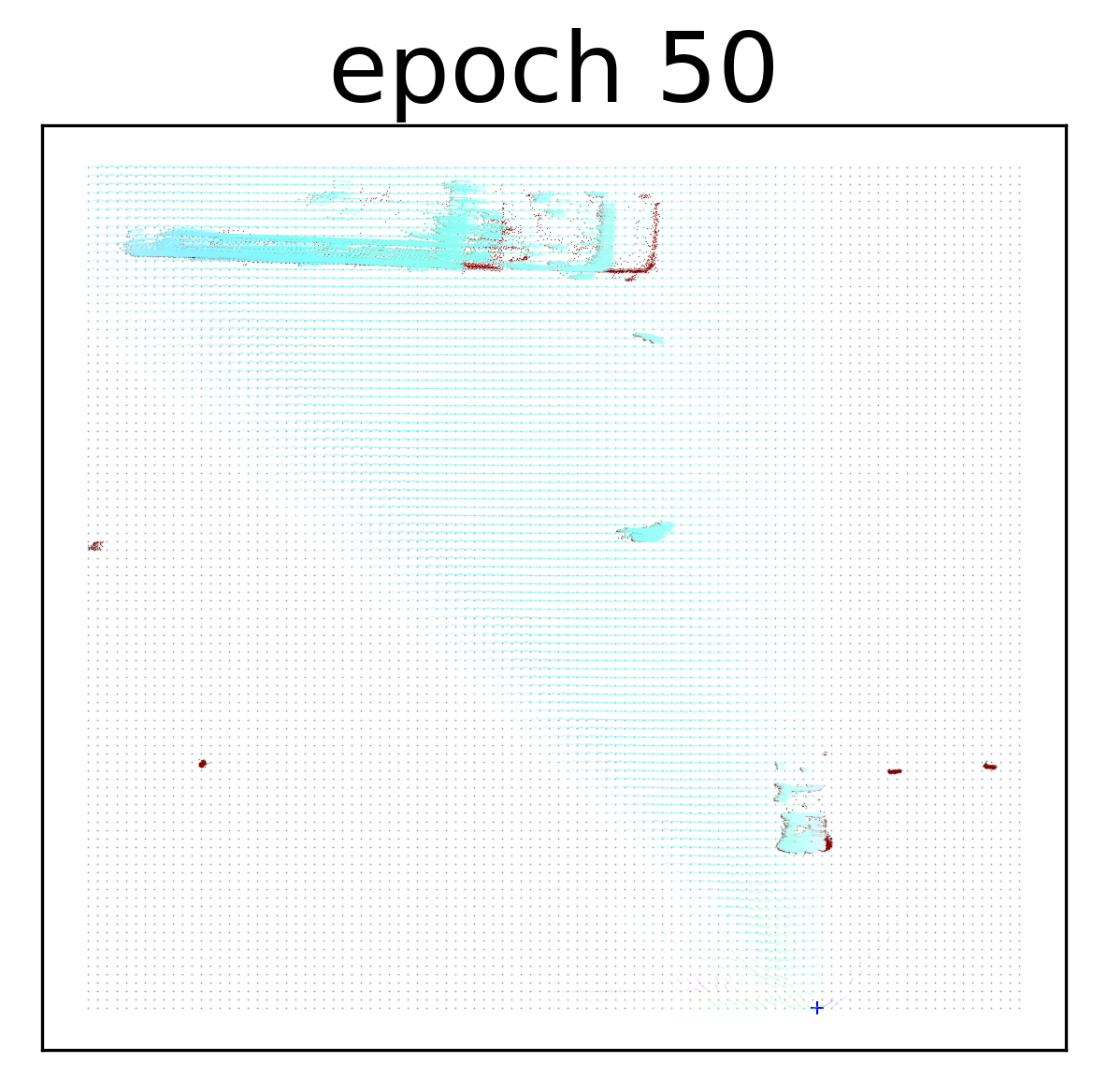}
        \includegraphics[width=0.32\textwidth,trim={0.39cm 0 0.25cm 0.25cm},clip]{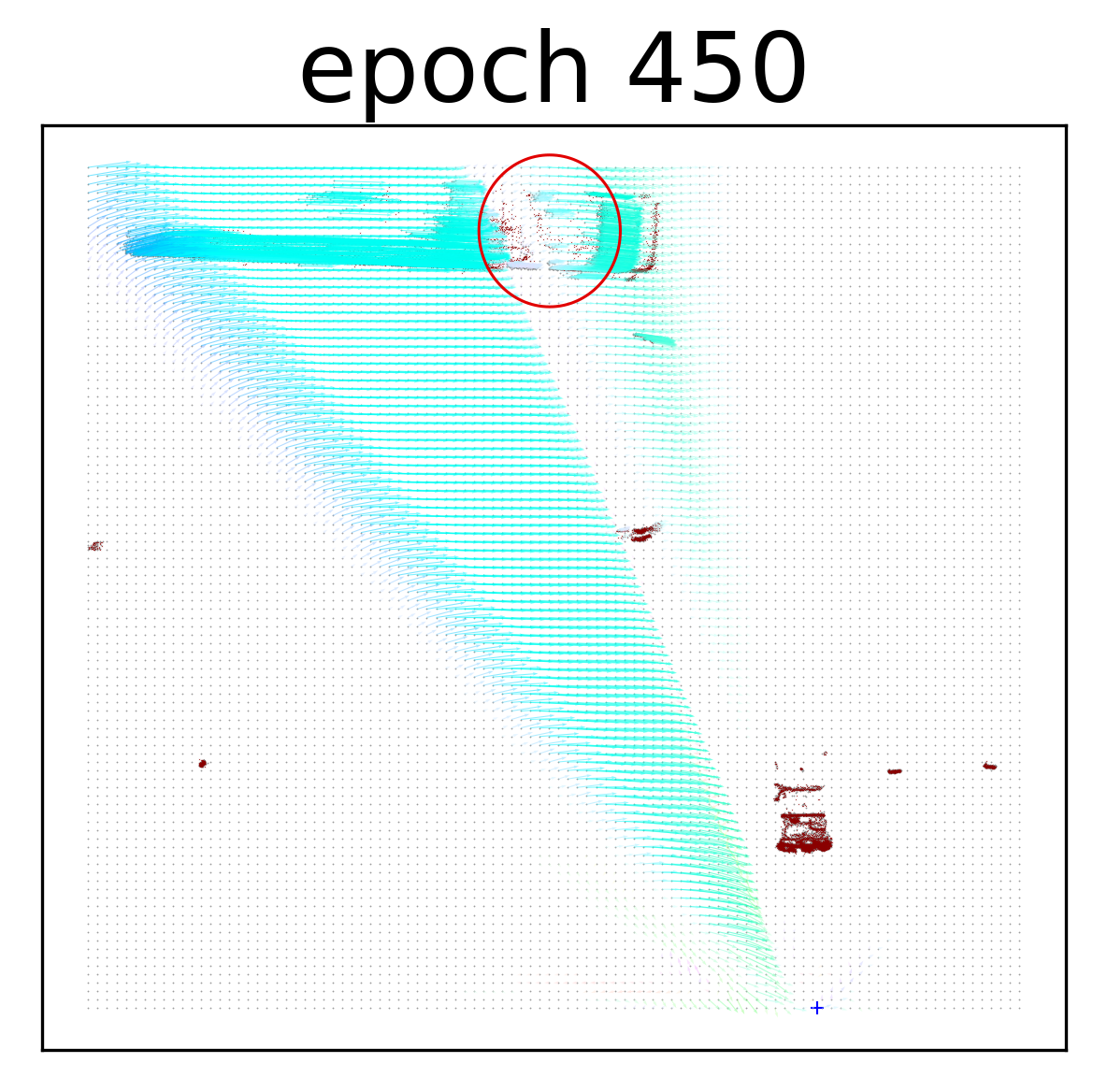}
        \caption{MLP}
    \end{subfigure}    
    \begin{subfigure}[b]{1.0\columnwidth}
        \centering
        \includegraphics[width=0.32\textwidth,trim={0.39cm 0 0.25cm 0.2cm}]{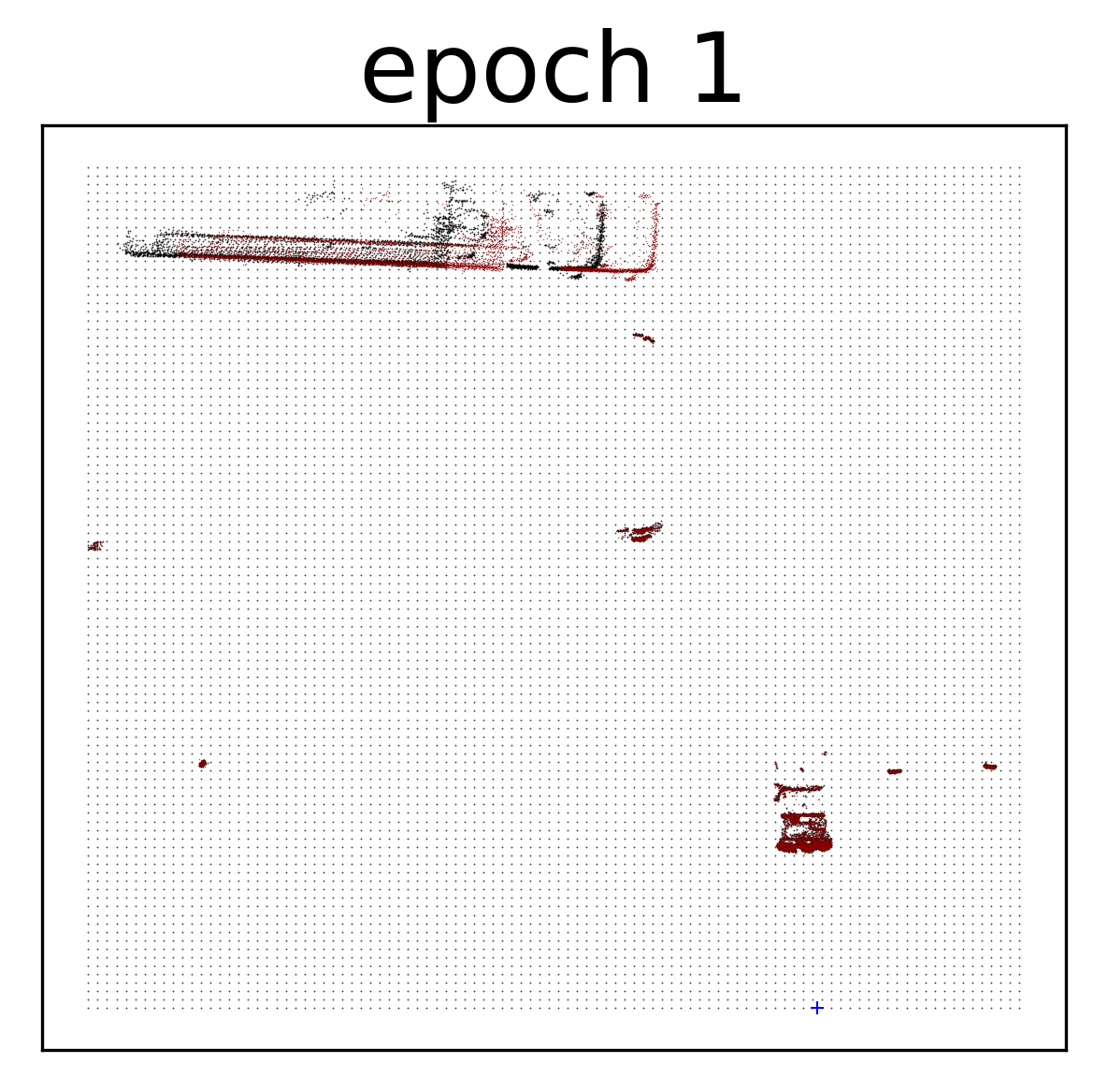}
        \includegraphics[width=0.32\textwidth,trim={0.39cm 0 0.25cm 0.2cm}]{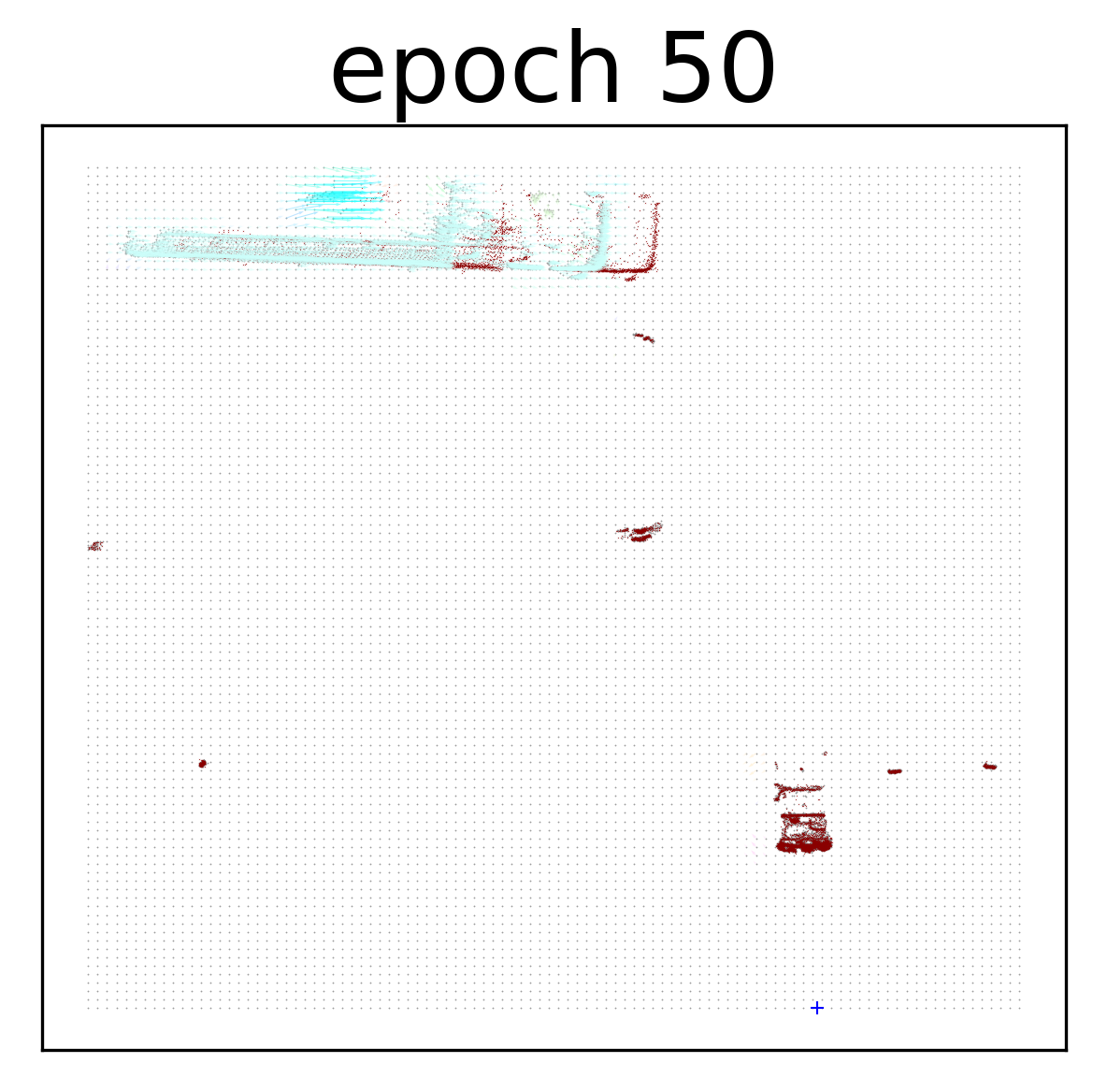}
        \includegraphics[width=0.32\textwidth,trim={0.39cm 0 0.25cm 0.2cm}]{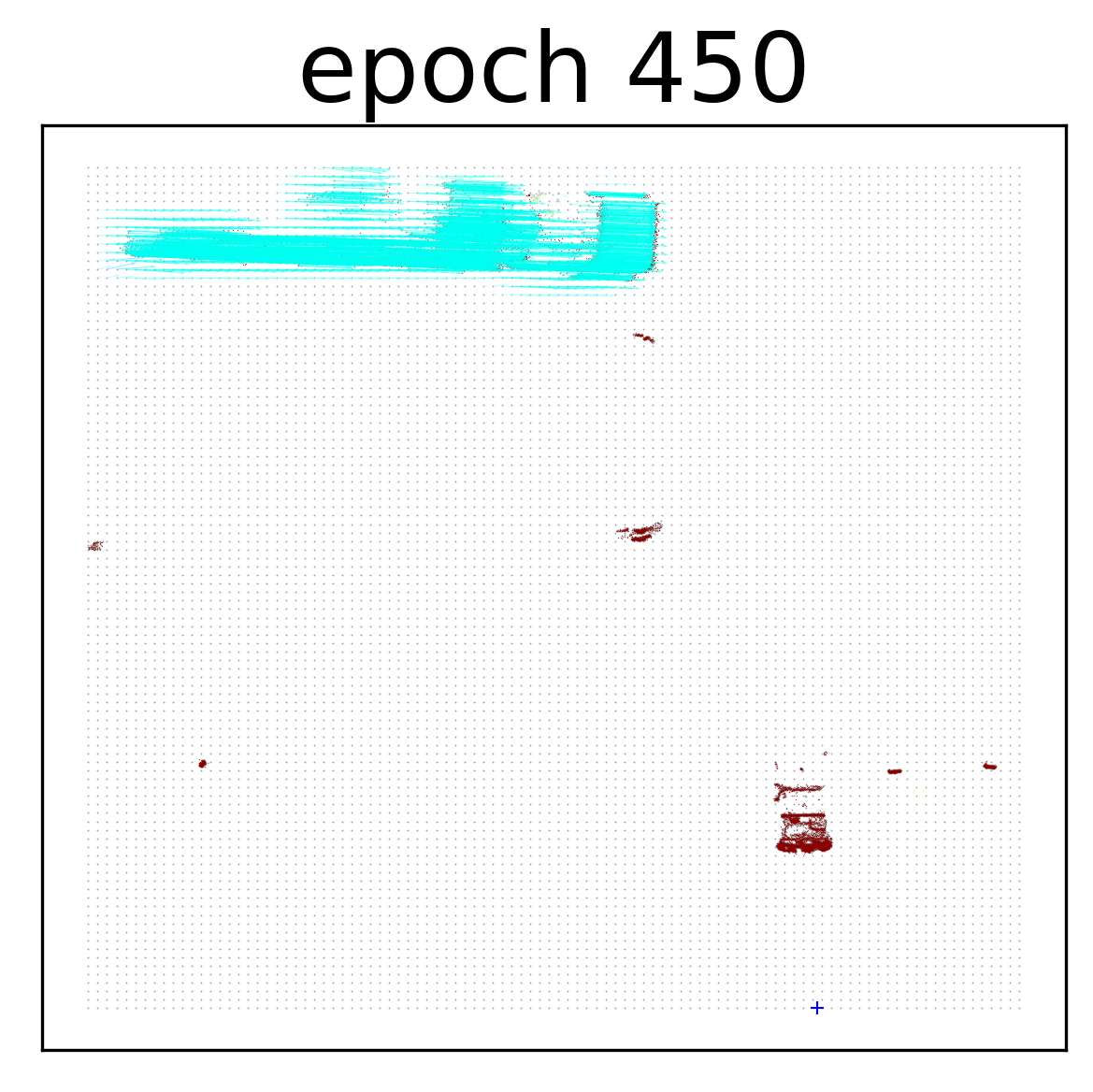}
        \caption{Floxels}
    \end{subfigure}
    \caption{\textbf{Evolution of estimated scene flow.} The birds-eye view of the flow during optimization. The MLP exhibits ``windmill artifacts'' throughout training and never unlearns them in some regions. The flow in {\color{RubineRed}shadow regions (red circle)} remains underestimated. Floxels starts with zero flow, learns flow only in regions with objects, gracefully overcomes the difficult shadow region, and converges faster to the true flow. Points at time $t$ are black and $t+1$ are red.
    Best seen on screen and zoomed in. A larger version is shown in the \cref{fig:flow_opt_vis_fnsf}.}
    \label{fig:flow_opt_vis_2}
\end{figure}
\begin{table}[h!]
    \caption{\textbf{Influence of different loss components.} Results obtained on nuScenes mini. All models use five scans. ``-'' indicates that the respective component got removed. We show the results for static points in \cref{tab:abl_loss_coponents_full}.}
    \label{tab:abl_loss_coponents}
    \centering
    \resizebox{\columnwidth}{!}{%
    \begin{tabular}{l|c c c c}
    \toprule
    Method & \multicolumn{4}{c}{Dynamic Points} \\
    \midrule
    & EPE $\downarrow$ & $\text{Acc}_{5}$ $\uparrow$ & $\text{Acc}_{10}$ $\uparrow$ & angle error \\
         \midrule
         \floxels & 0.085 & 0.537 & 0.833 & 0.489 \\
         ~- flow norm & 0.084 & 0.528 & 0.833 & 0.487 \\
         ~- cluster loss & 0.201 & 0.133 & 0.413 & 0.802 \\
         ~- cluster loss \&  & \multirow{2}{*}{0.206} & \multirow{2}{*}{0.123} & \multirow{2}{*}{0.401} & \multirow{2}{*}{0.793} \\
         ~~~- flow norm \\
         \bottomrule
    \end{tabular}
}    
\end{table}
\begin{table}[h]
    \centering
    \caption{\textbf{Influence of the number of scans.} Using nuScenes mini.}
    \label{tab:abl_nr_frames}    
    \resizebox{\columnwidth}{!}{
    \begin{tabular}{l|c c c c | c}
    \toprule
    Method & \multicolumn{4}{|c|}{Dynamic Points} \\
    \midrule
    & EPE $\downarrow$ & $\text{Acc}_{5}$ $\uparrow$ & $\text{Acc}_{10}$ $\uparrow$ & angle error $\downarrow$ &  Time (s) $\downarrow$ \\
         \midrule
        3 scans & 0.095 & 0.468 & 0.799 & 0.524 & \textbf{2.47} \\
        5 scans & 0.085 & \textbf{0.537} & 0.833 & 0.489 &  3.52\\
        7 scans & 0.082 & 0.533 & 0.839  & 0.474 &  4.61 \\
        9 scans & 0.078 & 0.516 & 0.852 & 0.460 &  5.69 \\
        11 scans & \textbf{0.076} & 0.486 & \textbf{0.864} & \textbf{0.447} & 6.72 \\
         \bottomrule
    \end{tabular}
 }  
\end{table}
\begin{figure*}[h!]
    \centering
    \begin{subfigure}[b]{.32\textwidth}
        \centering
        \includegraphics[width=1\textwidth,trim={0 0.5cm 0 3cm},clip]{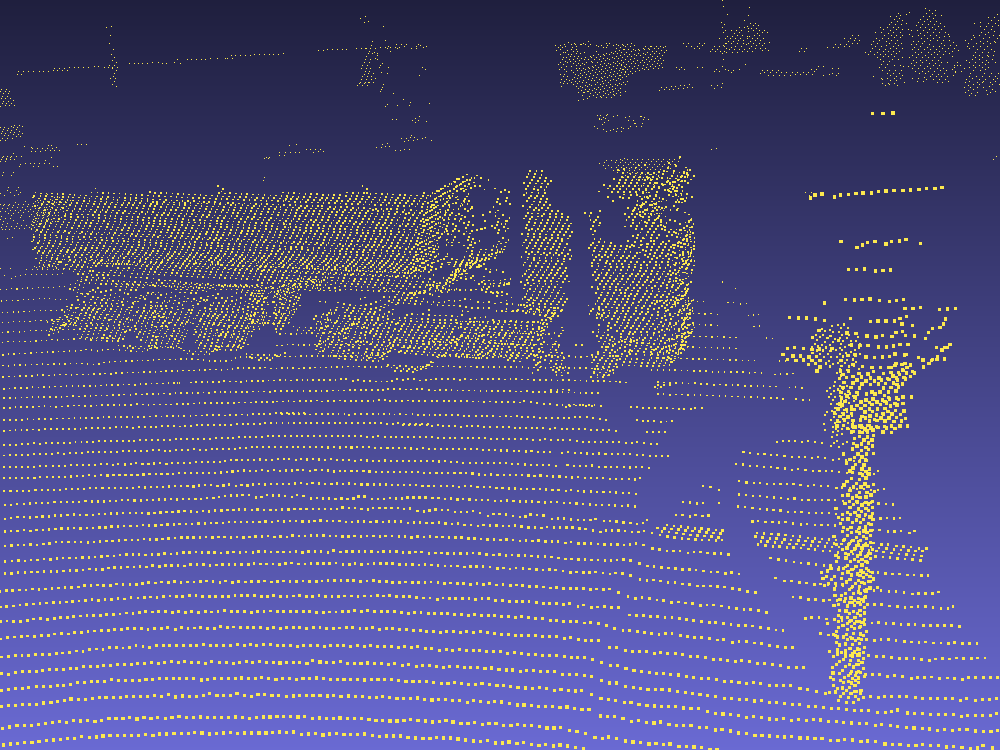}
        \caption{Single point cloud at $t_0$.}
    \end{subfigure}
    \begin{subfigure}[b]{.32\textwidth}
        \centering
        \includegraphics[width=1\textwidth,trim={0 0.5cm 0 3cm},clip]{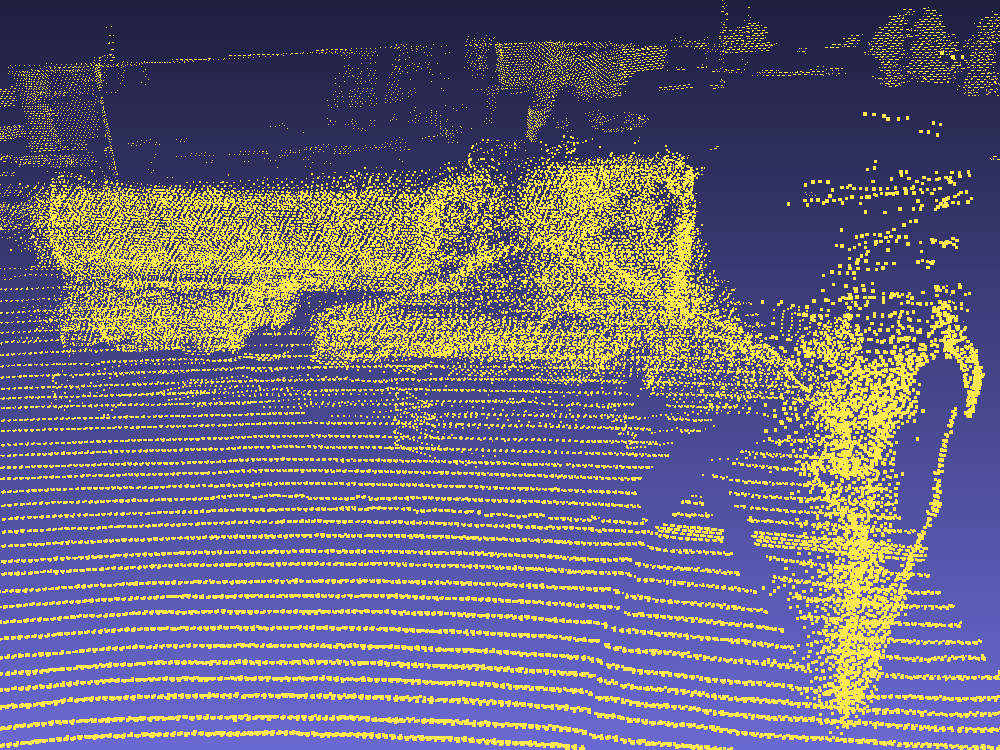}
        \caption{MLP used for point cloud accumulation.}
    \end{subfigure}    
    \begin{subfigure}[b]{.32\textwidth}
        \centering
        \includegraphics[width=1\textwidth,trim={0 0.5cm 0 3cm},clip]{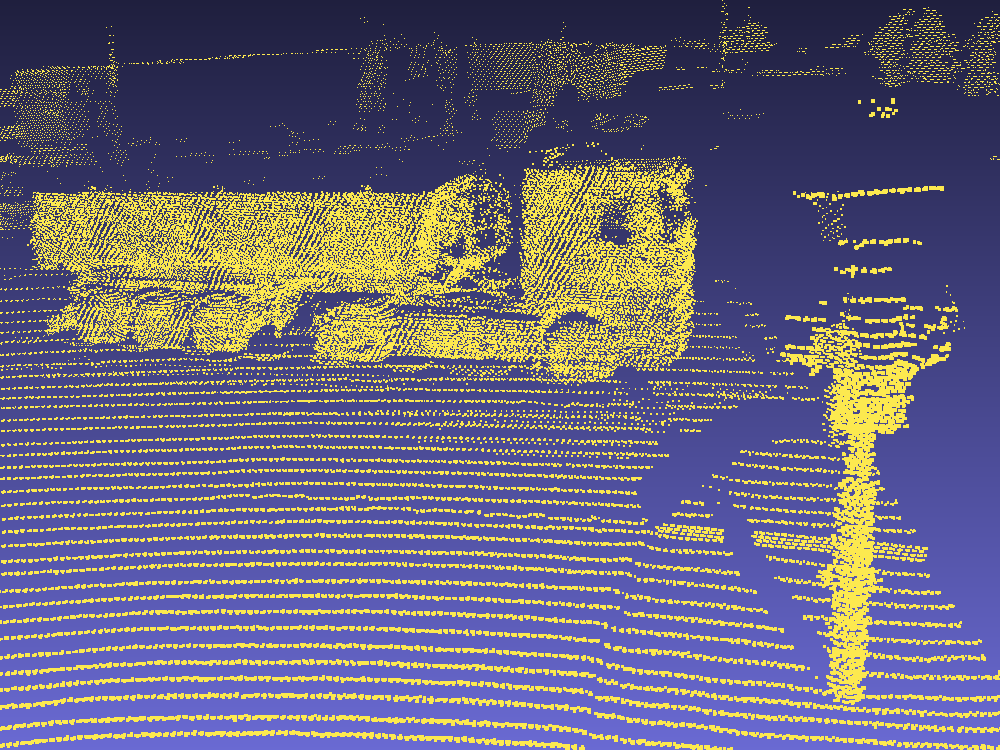}
        \caption{Floxels used for point cloud accumulation.}
    \end{subfigure}
    \caption{\textbf{Point cloud densification.} 
    Left: Single scan. Middle and Right: Point cloud accumulation using five adjacent point clouds. As can be seen, using an MLP leads to a noisy, unusable, dense point cloud. \floxels allow for precise, high-quality point cloud densification. 
    }
    \label{fig:point_cloud_densification}
\end{figure*}
\begin{table*}[h!]
    \caption{\textbf{Voxel size ablation -- Argoverse 2 validation split}. Varying voxel size has little effect on EPE across classes. A slight decrease in pedestrian performance can be observed for the largest voxel size (2~m). Larger voxel sizes slightly improve optimization speed.}
    \label{tab:bucket_val}
    \centering
    \begin{tabular}{c|c c c c c | c | c}
    \toprule
        \multicolumn{8}{c}{Argoverse 2 validation --- Static/Dynamic} \\
        \midrule
         Method & BG & Car & Other Vehicle & Pedestrian & Wheeled VRU & mdnEPE & Time (s) \\
         \midrule 
         Floxels 5 (0.3~m) & 0.021 & 0.022 / 0.103 & 0.014 / \textbf{0.358} & 0.019 / \textbf{0.281} & 0.017 / 0.123 & \textbf{0.216} & 4.46 \\
         Floxels 5 (0.5~m)  & 0.023 & 0.021 / 0.108 & 0.011 / 0.362 & 0.015 / 0.284 & 0.018 / \textbf{0.117} & 0.217 & 3.81\\ 
         Floxels 5 (0.7~m) & 0.019 & 0.022 / 0.114 & 0.013 / 0.366 & 0.012 / 0.282 & 0.019 / 0.118 & 0.220 & 3.71\\
         Floxels 5 (2.0~m) & 0.020 & 0.021 / \textbf{0.102} & 0.012 / 0.371 & 0.014 / 0.319 & 0.016 / 0.127 & 0.229 & 3.41 \\
         \bottomrule
    \end{tabular}
\end{table*}
\myparagraph{Runtime.}
All timings are measured on an Nvidia T4 GPU with 16~GB of RAM. Floxels 5 and Floxels 13 take $3.52$s and $9.28$s per frame, i.e., $\sim9$~min and $\sim24.13$~min for one sequence of 156 scans (\cref{tab:bucket_val}), compared to $\sim 24~h$ needed by EulerFlow on a V100 GPU \cite{vedder2024neuraleuleriansceneflow}.
Optimizing multiple Floxels in parallel can yield an additional 2-4$\times$ speedup on a T4.
Compared to NSFP and FNSP, runtime scales much better for \floxels with increasing point cloud size (compare \cref{tab:nusc_test_full,tab:argo}).  
On larger point clouds (\cref{tab:argo}), \floxels is $\sim4.7\times$ faster than FNSF and $\sim14.4\times$ faster than NSFP.
DifFlow3D \cite{liu2024difflow3drobustuncertaintyawarescene} is faster, 
but memory requirements prohibit its use on larger point clouds.

\subsection{Is the neural prior a good prior?}
\label{sec:neural_prior_good}
Previous work \cite{li2021neural} argued that using an MLP acts as a regularizer and a beneficial prior. 
We observe, that the neural prior tends to create overly smooth scene flow fields that extend far beyond the actual moving objects. This is especially apparent in less busy scenes with few moving objects and empty regions.
As shown in Fig.~\ref{fig:birds_eye_flow}, the scene flow field displays windmill artifacts, as the neural prior extends flow fields along the rays with origin at the sensor location.
In contrast, the scene flow of \floxels is more localized around moving objects and does not display similar artifacts. 
Note, that the windmill artifacts lead to wrong lidar-to-camera synchronization (Fig.~\ref{fig:flow_opt_vis_2}).
Fig.~\ref{fig:birds_eye_flow} also shows that even small occlusions, here caused by the traffic light, lead to zero-flow for the occluded region of the truck, while \floxels predicts the correct flow.

\subsection{Qualitative results}
\label{sec:qualit_results}
Here, we compare the properties of MLP-based scene flow estimation and voxel grid-based estimation.

\myparagraph{Evolution of estimated scene flow field.}
Fig.~\ref{fig:flow_opt_vis_2} shows the evolution of the flow field as the optimization progresses. The MLP displays strong windmill artifacts early on, showing that it struggles to learn good localized spatial representations, i.e.~moving points influence the flow field in far regions, and small objects are initially neglected. Both slow down convergence.
In contrast, the gradients for \floxels have only local influence on a few voxels. Thus, they do not show similar artifacts. 
Even the flow of small objects can be learned from the start of the optimization.

\myparagraph{Point cloud densification.}
Due to cubic space scaling, lidar point clouds become sparse at greater distances from the sensor, limiting a variety of tasks. A common solution involves accumulating points from multiple clouds using estimated scene flow.
MLP-based methods can fail at the task, especially in corner cases like occluded objects (Fig.~\ref{fig:point_cloud_densification}).
\floxels generates high-quality densified point clouds, revealing fine details of the truck. This showcases the precise flow field learned by \floxels.

\subsection{Influence of different components}
To evaluate the different components, we run \floxels and ablate its components. We summarize the results in Tab.~\ref{tab:abl_loss_coponents}.
Ablating the flow norm loss leads to a minor decrease for both dynamic and static points. Removing the cluster loss leads to significantly worse results, which are comparable to FNSF.
Removing both the flow norm and cluster loss causes accuracy to drop further.

\myparagraph{Influence of the number of scans }
on the runtime and accuracy. 
As can be seen in Tab.~\ref{tab:abl_nr_frames}, the runtime increases with each scan pair by roughly one second. However, improvements start to diminish (for Argoverse 1 and nuScenes) beyond 5 scans, so we use 5 as our base setting for a good tradeoff between runtime and performance.

\myparagraph{Voxel size.}
Finally, we evaluate the influence of the voxel size on both performance and runtime on Argoverse 2. \cref{tab:bucket_val} shows that a smaller voxel size (0.3m) leads to slightly better results on average, but comes with higher inference cost.
We pick 0.5 for a good tradeoff between performance and runtime.
\section{Conclusion}
In this work, we have investigated test-time optimization methods for scene flow estimation on ego-motion corrected data.
This investigation revealed severe shortcomings of existing methods, like non-local influence of flow estimates, large flow predictions in empty regions, and difficulties with occlusion.
As a remedy, we proposed using a simple voxel grid as a model and additional loss terms.
We showed, qualitatively and quantitatively, that our method \floxels handles these challenging cases significantly better.
\floxels not only outperforms previous test-time optimization methods by a large margin but also runs and converges faster and scales well to large point clouds. While the concurrent work EulerFlow is slightly better, \floxels is $\sim 60-140\times$ faster.

{
    \small
    \bibliographystyle{ieeenat_fullname}
    \bibliography{egbib}
}

\clearpage
\newpage
\appendix
\section{Supplemental material}

\subsection{Details on nuScenes and Argoverse 1 protocol}
\label{sec:data_generation_details}
\paragraph{Preprocessing.}
We subsampled the datasets to 10 HZ with a tolerance of 0.1 HZ. We discard all scan sequences that violate this tolerance to ensure a high quality.
While the datasets provide annotations for all, only a subset of frames was labeled by humans. For all remaining frames labeles were generated automatically. 
To ensure the highest possible quality, we evaluate only on scan sequences where at least one of the frames was annotated by a human. 
To ignore the ego vehicle, we discard all points within a radius of 3 meters around the origin. We also discard points higher than 4 meters and further away than 50 meters, as the point clouds become very sparse, which can result in noisy pseudo flow.

Before creating the pseudo ground truth flow, we discard points belonging to the ground plane; as for these points, the flow is commonly noisy (due to the circular scanning patterns on the ground plane).
We discard ground points by estimating the ground plane using Progressive Morphological Filtering (PMF) \cite{zhang2003progressive}.
PMF applies a series of filtering steps and progressively refines the ground plane estimate. 

To compensate for the ego-motion, we use the remaining points after discarding the ground points and filter all points labeled as potentially dynamic points in the datasets.
The remaining points are neither dynamic nor do they belong to the ground.
We use these points to estimate a transformation between the point clouds by applying KISS-ICP \cite{Vizzo_2023}.
We apply this transformation to all non-ground points in the point cloud (including the dynamic points) to align the static parts of the two point clouds. 
After this transform, we compute the flow for each point as described below.

\paragraph{Ground truth scene flow for 3D objects.}
After compensating for the ego-motion described above, we use the bounding box annotations of the datasets to estimate the flow of the dynamic objects. We compute the transformation from each bounding box to the corresponding annotation in the next frame and use the corresponding transformation to calculate the ground-truth flow of all points within the 3D bounding boxes. 
The annotated objects might not be moving but can be static, e.g., a parked car.
To automatically label points as ``static'' or ``dynamic'' we compute the mean motion over all points of each potentially dynamic object. If this motion is larger than $0.5~\mathrm{m/s}$  we label them as ``dynamic''.

\subsection{Baseline details and hyperparameters}
\label{sec:sup_baseline_detail}
\paragraph{Early stopping.}
For all experiments with NSFP and FNSF with early stopping activated, we keep the default parameters as published by the authors\footnote{\url{https://github.com/Lilac-Lee/Neural_Scene_Flow_Prior}}$^,$\footnote{\url{https://github.com/Lilac-Lee/FastNSF}}. We set the early stopping patience to 100 epochs and early stopping minimum delta to 0.0001. The maximum number of epochs is set to 5000.

\paragraph{Learning rate.}
All baseline models are trained with the Adam optimizer. We set the learning rate for FNSF-8 to its default value of 0.001. In the case of NSFP-8 and NSFP-16, we diverge from the default value and set the learning rate to 0.0008. Finally, staying in line with the default settings, we do not use weight decay.

\paragraph{MLP hidden units.}
For all baseline experiments, we keep the number of hidden units to its default value of 128.

\subsection{\floxels details and hyperparameters}

\paragraph{Early stopping.}
For experiments with Floxels, we train for a maximum of 500 epochs, set the early patience to 250 epochs, and set the early stopping minimum delta to 0.01. 
Whenever it is stated that early stopping is identical to the baselines, we use the early stopping as described in Sec.~\ref{sec:sup_baseline_detail} and set the maximum epochs to 5000 for consistency.

\subsection{Complete quantitative results}
\begin{table*}[t]
    \caption{\textbf{Static/Dynamic Normalized EPE on Argoverse 2 (2024) Scene Flow Challenge test set} \cite{khatri2025can}. Baseline scores from challenge leaderboard.}
    \label{tab:bucket_val_full}
    \centering
    \begin{tabular}{l l|c c c c c | c}
    \toprule
         & Method & BG & car & other vehicle & pedestrian & wheeled VRU & mDEPE \\
    \toprule
    \multirow{3}{*}{\rotatebox{90}{Superv.}} 
    & Flow4D \cite{kim2024flow4dleveraging4dvoxel}
      & 0.005 & 0.087 & 0.150 & 0.216 & 0.127 & 0.145 \\
    & TrackFlow \cite{khatri2025can}
      & 0.002 & 0.182 & 0.305 & 0.358 & 0.230 & 0.269 \\
    & DeFlow \cite{zhang2024deflowdecodersceneflow}
      & 0.005 & 0.113 & 0.228 & 0.496 & 0.266 & 0.276 \\
    \midrule
    \multirow{8}{*}{\rotatebox{90}{Unsupervised}}
    & NSFP \cite{li2021neural}
      & 0.034 & 0.251 & 0.331 & 0.722 & 0.383 & 0.422 \\
    & Fast NSF \cite{li2023fast}
      & 0.091 & 0.296 & 0.413 & 0.500 & 0.322 & 0.383 \\
    & Zeroflow XL 5x \cite{vedder2024zeroflowscalablesceneflow}
      & 0.013 & 0.238 & 0.258 & 0.808 & 0.452 & 0.439 \\
    & Liu et al. 2024 \cite{liu2024selfsupervisedmultiframeneuralscene}
      & 0.106 & 0.310 & 0.559 & 0.509 & 0.276 & 0.413 \\
    & SeFlow \cite{zhang2024seflowselfsupervisedsceneflow}
      & 0.006 & 0.214 & 0.291 & 0.464 & 0.265 & 0.309 \\      
    & Euler Flow \cite{vedder2024neuraleuleriansceneflow}
      & 0.053 & 0.093 & 0.141 & 0.195 & 0.093 & 0.130 \\
    & Floxels 5 0.5~m (ours)
      & 0.024 & 0.119 & 0.194 & 0.243 & 0.113 & 0.168 \\
    & Floxels 9 0.5~m (ours)
      & 0.018 & 0.108 & 0.202 & 0.208 & 0.100 & 0.155 \\
    & Floxels 13 0.5~m (ours)
      & 0.015 & 0.112 & 0.213 & 0.195 & 0.096 & 0.154 \\
    \bottomrule
    \end{tabular}
\end{table*}

\begin{table*}[ht]
    \centering
    \caption{\textbf{Results on nuScenes validation set.} Models trained without early stopping (5000 epochs) denoted with “*”. “-N” indicates the number of layers. For a fair comparison we provide timings only when using early stopping.}
    \label{tab:nusc_test_full}    
    \begin{tabular}{l|c c c c | c c c | c} 
    \toprule
    Method & \multicolumn{4}{|c|}{Dynamic Points} & \multicolumn{3}{|c}{Static Points} \\
    \midrule
    & EPE $\downarrow$ & $\text{Acc}_{5}$ $\uparrow$ & $\text{Acc}_{10}$ $\uparrow$ & angle error $\downarrow$ & EPE $\downarrow$ & $\text{Acc}_{5} $ $\uparrow$ & $\text{Acc}_{10}$ $\uparrow$ & Time (s) \\ 
        \midrule
         & \multicolumn{7}{c}{Supervised} \\
        \midrule
        DifFlow3D & 0.089 & 0.554 &  0.823 & 0.325 & 0.044 & 0.748 & 0.935 & 0.48\\
         \midrule
         & \multicolumn{7}{c}{Self-supervised test-time optimization} \\
         \midrule
         NSFP-8  & 0.141 & 0.316 & 0.636 & 0.471  & 0.068 & 0.613 & 0.861 & \textbf{3.43}\\
         NSFP-8* & 0.139 & 0.315 & 0.641 & 0.470  & 0.067 & 0.613 & 0.861 & --\\
         NSFP-16 & 0.148 & 0.322 & 0.647 & 0.488  & \textbf{0.061} & 0.664 & \textbf{0.883} & 8.99\\
         NSFP-16* & 0.145 & 0.384 & 0.679 & 0.460  & 0.087 & 0.619 & 0.818 & --\\
         FNSF-8 & 0.266 & 0.211 & 0.501 & 0.628 & 0.137  & 0.439 & 0.723 & 5.93\\
         FNSF-8* & 0.372 & 0.122 & 0.361 & 0.757  & 0.241 & 0.241 & 0.531 & --\\
         \floxels\! (5s) & \textbf{0.102} & \textbf{0.464} & \textbf{0.786} & \textbf{0.430}  & 0.063 & \textbf{0.755} & 0.881 & 4.35\\
         \bottomrule
    \end{tabular}
\end{table*}

\begin{table*}[ht]
    \centering
    \caption{\textbf{Results on Argoverse test set.} Models trained without early stopping (5000 epochs) denoted with “*”. “-N” indicates the number of layers.}
    \label{tab:argo_full}
    \begin{tabular}{l|c c c c | c c c | c}
    \toprule
    Method & \multicolumn{4}{|c|}{Dynamic Points} & \multicolumn{3}{|c}{Static Points} & \multicolumn{1}{|c}{Time (s)}\\
    \midrule
    & EPE $\downarrow$ & $\text{Acc}_{5}$ $\uparrow$ & $\text{Acc}_{10}$ $\uparrow$ & angle error $\downarrow$ & EPE $\downarrow$ & $\text{Acc}_{5}$ $\uparrow$ & $\text{Acc}_{10}$ $\uparrow$ \\ 
        \midrule
        & \multicolumn{7}{c}{Supervised} \\
        \midrule
        DifFlow3D & \multicolumn{7}{c}{Out of Memory} \\
         \midrule
         & \multicolumn{7}{c}{Test-time optimization with same early stopping} \\
         \midrule
         NSFP-8 & 0.200 & 0.288 & 0.521 & 0.468 & 0.046 & 0.745 & 0.933 & 63.01 \\
         NSFP-16 & 0.226 & 0.280 & 0.498 & 0.530 & 0.045 & 0.772 & 0.942 & 72.54 \\
         FNSF-8 & 0.282 & 0.281 & 0.518 & 0.588 & 0.065 & 0.704 & 0.905 & 20.66 \\
         \floxels (5s) & \textbf{0.104} & \textbf{0.537} & \textbf{0.755} & \textbf{0.420} & \textbf{0.024} & \textbf{0.919} & \textbf{0.962} & \textbf{4.38} \\
          \midrule
         & \multicolumn{7}{c}{Test-time optimization.} \\
         \midrule
         NSFP-8* & 0.202 & 0.272 & 0.508 & 0.478 & 0.047 & 0.743 & 0.931 & - \\
         NSFP-16* & 0.203 & 0.336 & 0.541 & 0.495 & 0.043 & 0.815 & 0.948 & - \\
         FNSF-8* & 0.370 & 0.215 & 0.458 & 0.651 & 0.148 & 0.463 & 0.723 & - \\
         \floxels (5s) & \textbf{0.109} & \textbf{0.526} & \textbf{0.739} & \textbf{0.423} & \textbf{0.024} & \textbf{0.912} & \textbf{0.962} & - \\
         \bottomrule
    \end{tabular}
\end{table*}

\begin{table*}[ht]
    \centering
    \caption{\textbf{Influence of the number of scans.} Using nuScenes mini.}
    \label{tab:abl_nr_frames_full}
    \begin{tabular}{l|c c c c | c c c | c}
    \toprule
    Method & \multicolumn{4}{|c|}{Dynamic} & \multicolumn{3}{|c}{Static} \\
    \midrule
    & EPE $\downarrow$ & $\text{Acc}_{5}$ $\uparrow$ & $\text{Acc}_{10}$ $\uparrow$ & angle error $\downarrow$ & EPE $\downarrow$ & $\text{Acc}_{5}$ $\uparrow$ & $\text{Acc}_{10}$ $\uparrow$ &  Time (s) $\downarrow$ \\
         \midrule
        3 scans & 0.095 & 0.468 & 0.799 & 0.524 & 0.064 & 0.719 & 0.887 & \textbf{2.47} \\
        5 scans & 0.085 & \textbf{0.537} & 0.833 & 0.489 & 0.057 & 0.797 & 0.909 &  3.52\\
        7 scans & 0.082 & 0.533 & 0.839  & 0.474 & 0.051 & 0.816 & 0.916  & 4.61 \\
        9 scans & 0.078 & 0.516 & 0.852 & 0.460 & 0.048 & 0.830 & 0.923 & 5.69 \\
        11 scans & \textbf{0.076} & 0.486 & \textbf{0.864} & \textbf{0.447} & \textbf{0.045} & \textbf{0.840} & \textbf{0.929} & 6.72 \\
         \bottomrule
    \end{tabular}
\end{table*}

\begin{table*}[ht]
    \centering
    \caption{\textbf{Influence of different loss components.} Results obtained on nuScenes mini. All models use five scans. “-” indicates that the respective component got removed.}
    \label{tab:abl_loss_coponents_full}
    \begin{tabular}{l|c c c c | c c c c}
    \toprule
    Method & \multicolumn{4}{c|}{Dynamic Points} & \multicolumn{3}{|c}{Static Points} \\
    \midrule
    & EPE $\downarrow$ & $\text{Acc}_{5}$ $\uparrow$ & $\text{Acc}_{10}$ $\uparrow$ & angle error $\downarrow$ & EPE $\downarrow$ & $\text{Acc}_{5}$ $\uparrow$ & $\text{Acc}_{10}$ $\uparrow$ \\
         \midrule
         \floxels & 0.085 & 0.537 & 0.833 & 0.489 & 0.057 & 0.797 & 0.909\\
         ~- flow norm & 0.084 & 0.528 & 0.833 & 0.487 & 0.069 & 0.735 & 0.879\\
         ~- cluster loss & 0.201 & 0.133 & 0.413 & 0.802 & 0.153 & 0.205 & 0.493 \\
         ~- cluster loss and - flow norm & 0.206 & 0.123 & 0.401 & 0.793 & 0.182 & 0.153 & 0.420 & \\        
         \bottomrule
    \end{tabular}
\end{table*}

In the experiments section, we omit results for static points. Tables \ref{tab:nusc_test_full}, \ref{tab:argo_full}, \ref{tab:abl_nr_frames_full}, and \ref{tab:abl_loss_coponents_full} show the full results.
The official leaderboard for Argoverse 2 (2024) Scene Flow Challenge can be found here\footnote{\url{https://eval.ai/web/challenges/challenge-page/2210/leaderboard/5463}}.

\subsection{Qualitative comparison of FNSF, FNSF with Floxel losses and \floxels}
\begin{figure*}[t]
    \centering
    \begin{subfigure}[b]{1.0\textwidth}
        \centering
        \includegraphics[width=0.60\textwidth]{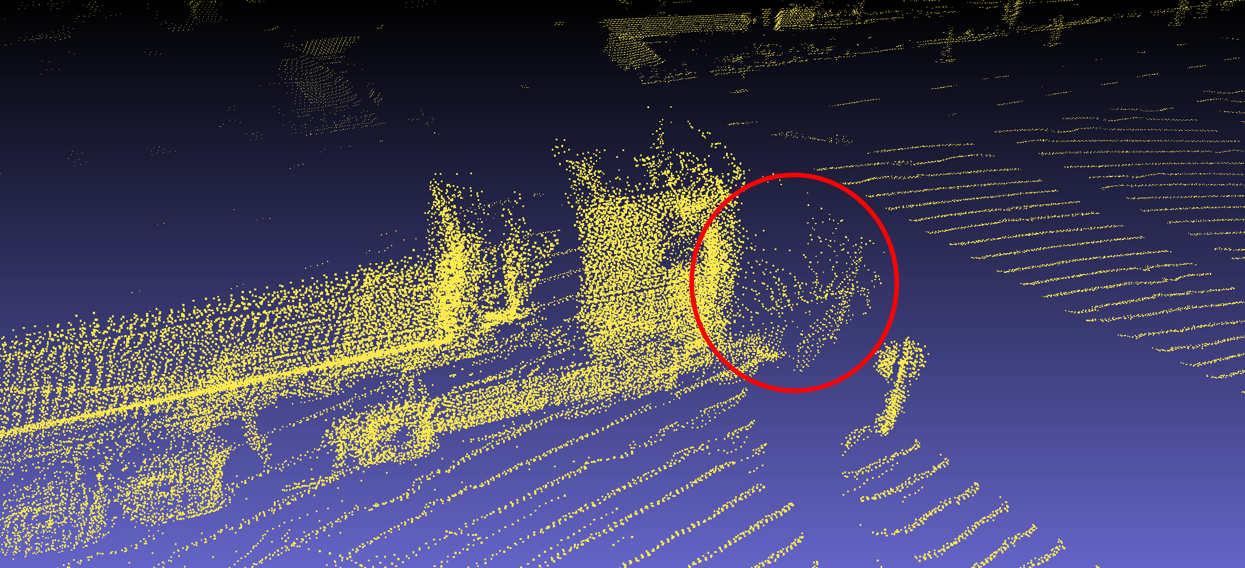}
        
        \caption{Original FNSF}
    \end{subfigure}
    \begin{subfigure}[b]{1.0\textwidth}
        \centering
        \includegraphics[width=0.60\textwidth]{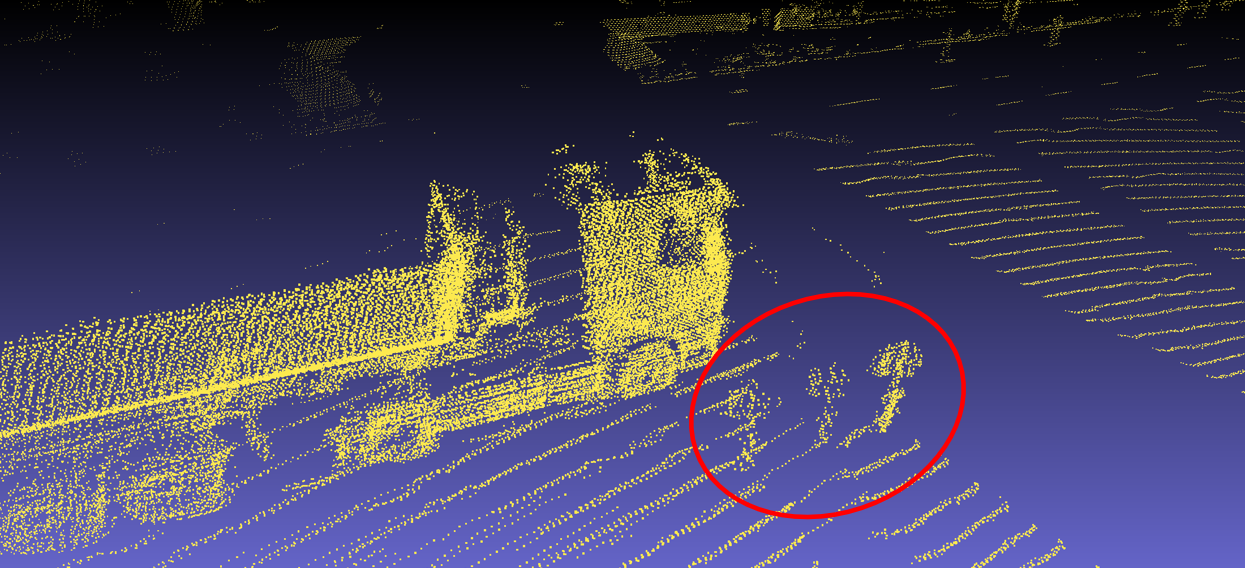}
        \caption{FNSF MLP with Floxels losses}
    \end{subfigure}
    \begin{subfigure}[b]{1.0\textwidth}
        \centering
        \includegraphics[width=0.60\textwidth]{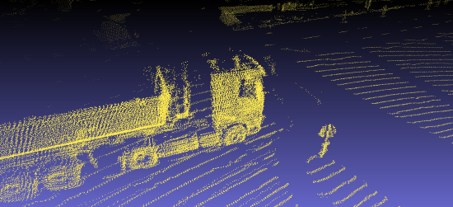}
        \caption{Floxels}
    \end{subfigure}
    \caption{\textbf{Accumulation over time compared between FNSF, FNSF with \floxels losses and \floxels}. We accumulate five point clouds t-2, t-1, t, t+1, and t+2. For FNSF parts of the front of the truck are moved too far forward. For FSNF MLP with Floxels loss the traffic sign is falsely affected by the scene flow field, which makes it appear 3 times in the accumulated point cloud. Floxes shows a much cleaner accumulated point cloud with more details.
    }

    \label{fig:accumulation_fnsf}
\end{figure*}

For the qualitative results in Sec.~\ref{sec:neural_prior_good} and Sec.~\ref{sec:qualit_results}, we use a slight variation of FNSF. In particular, we extended the losses of FNSF with a rigidity loss (similar to our clustering loss) and trained an MLP with 16 instead of 8 layers. 
We use this variant as our primary qualitative baseline as it performs on average better on our qualitative comparison dataset.
For completeness, we show the qualitative results for the original FNSF in Fig.~\ref{fig:flow_opt_vis_fnsf} and Fig.~\ref{fig:accumulation_fnsf}.

\begin{figure*}[t]
    \centering
    \begin{subfigure}[b]{0.42\textwidth}
        \centering
        \includegraphics[width=0.9\textwidth]{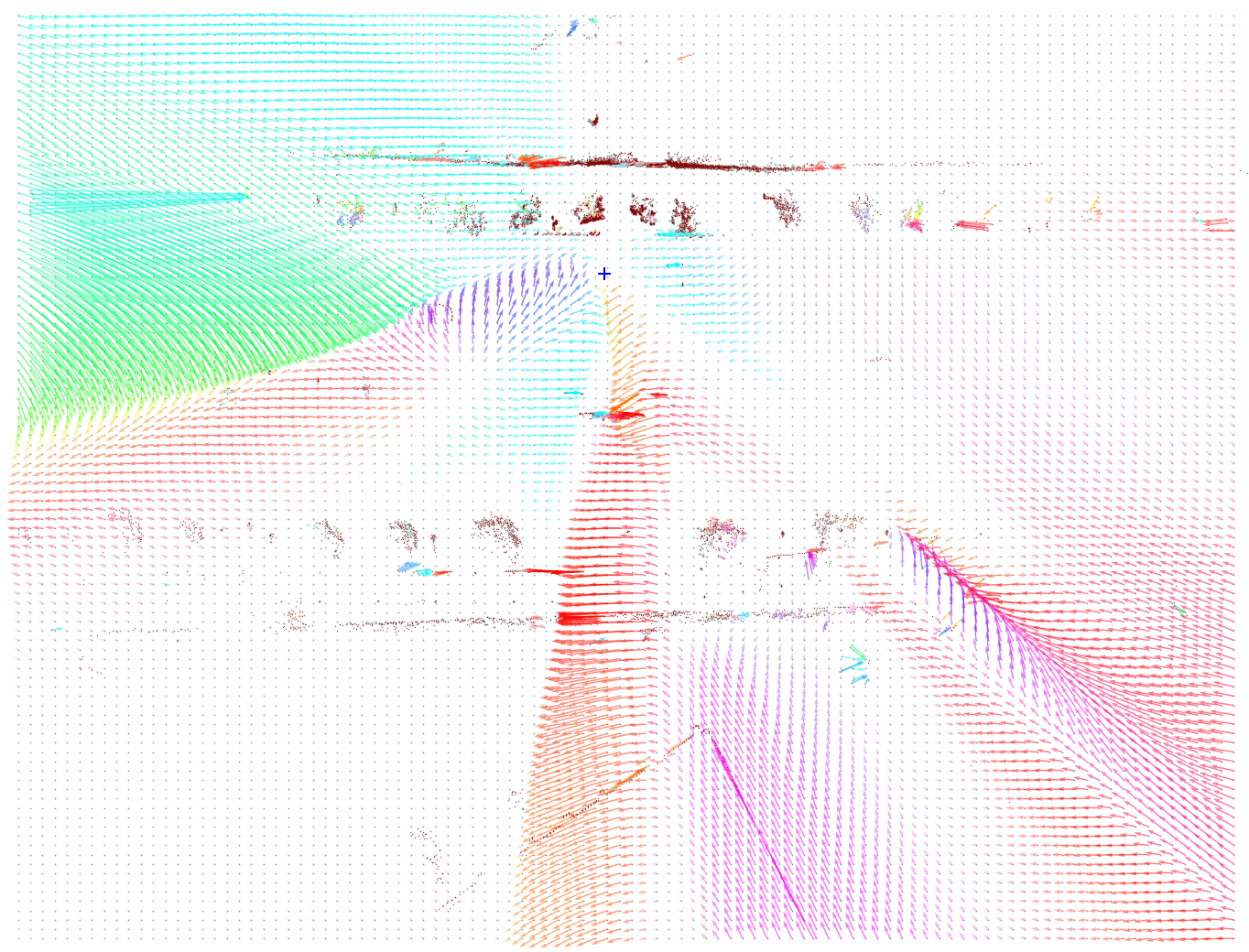}
    \end{subfigure}
    \begin{subfigure}[b]{0.42\textwidth}    
        \includegraphics[width=0.9\textwidth]{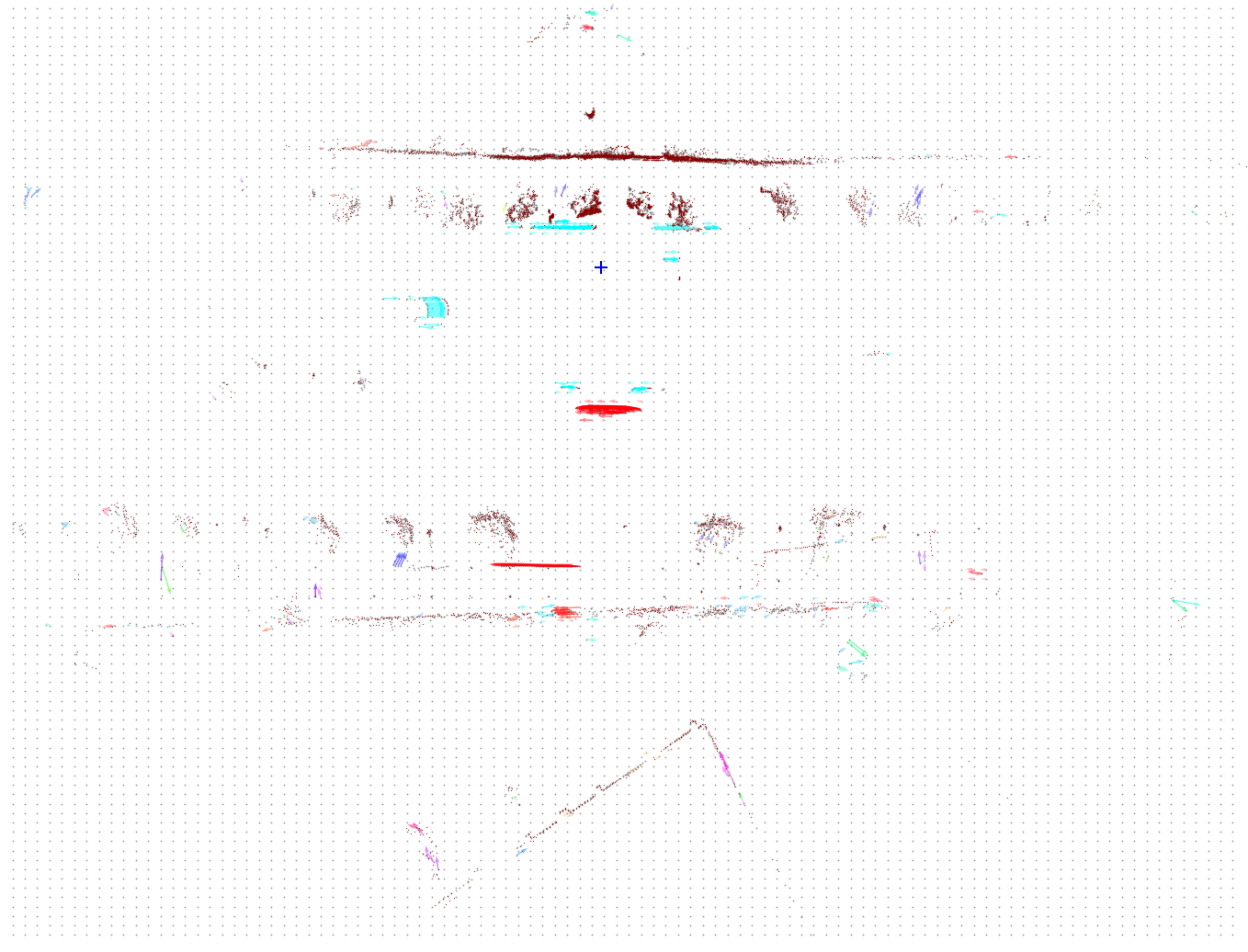}
    \end{subfigure}
    \begin{subfigure}[b]{0.42\textwidth}    
        \centering
        \includegraphics[width=0.9\textwidth]{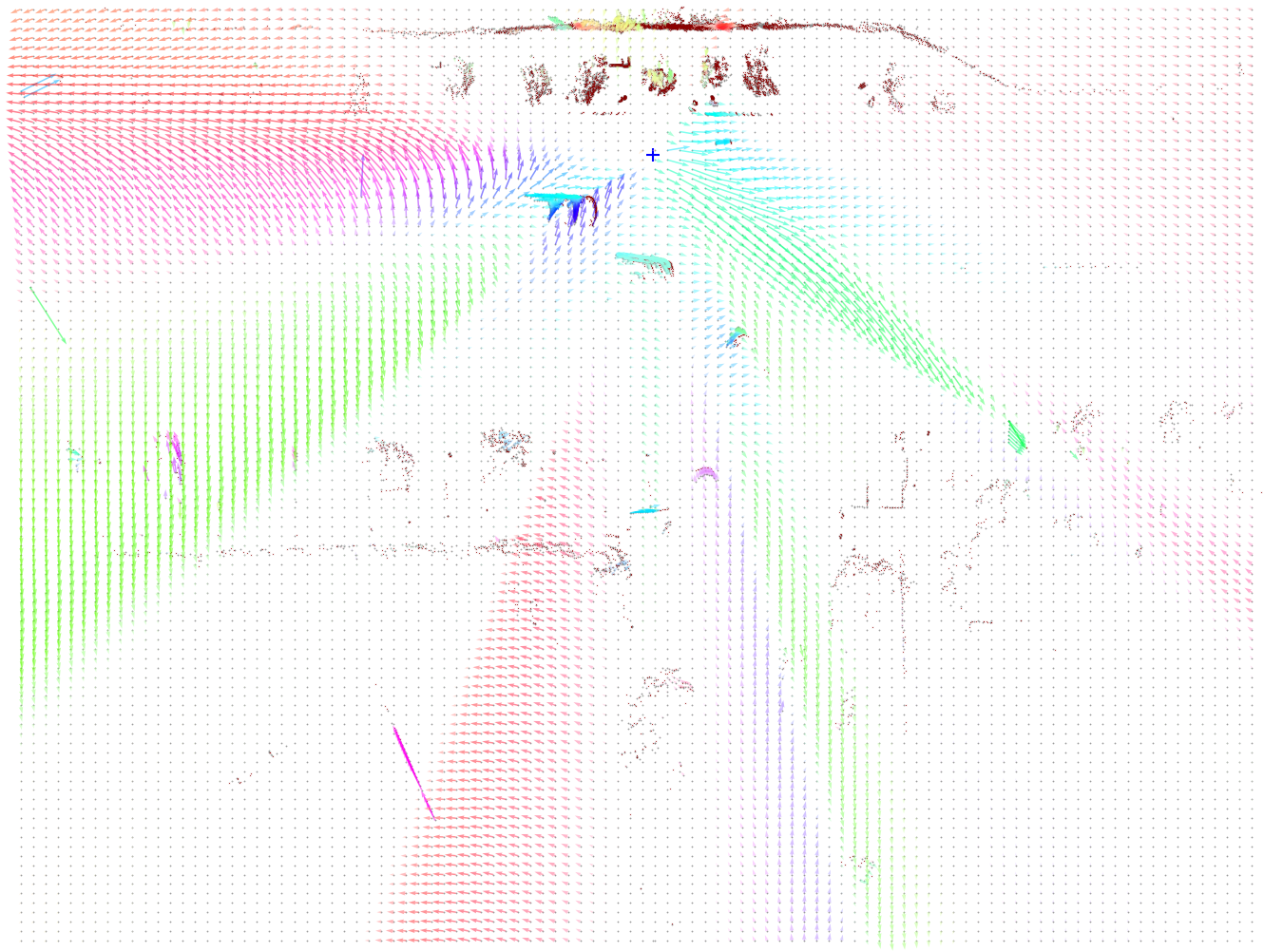}
    \end{subfigure}
    \begin{subfigure}[b]{0.42\textwidth}    
        \includegraphics[width=0.9\textwidth]{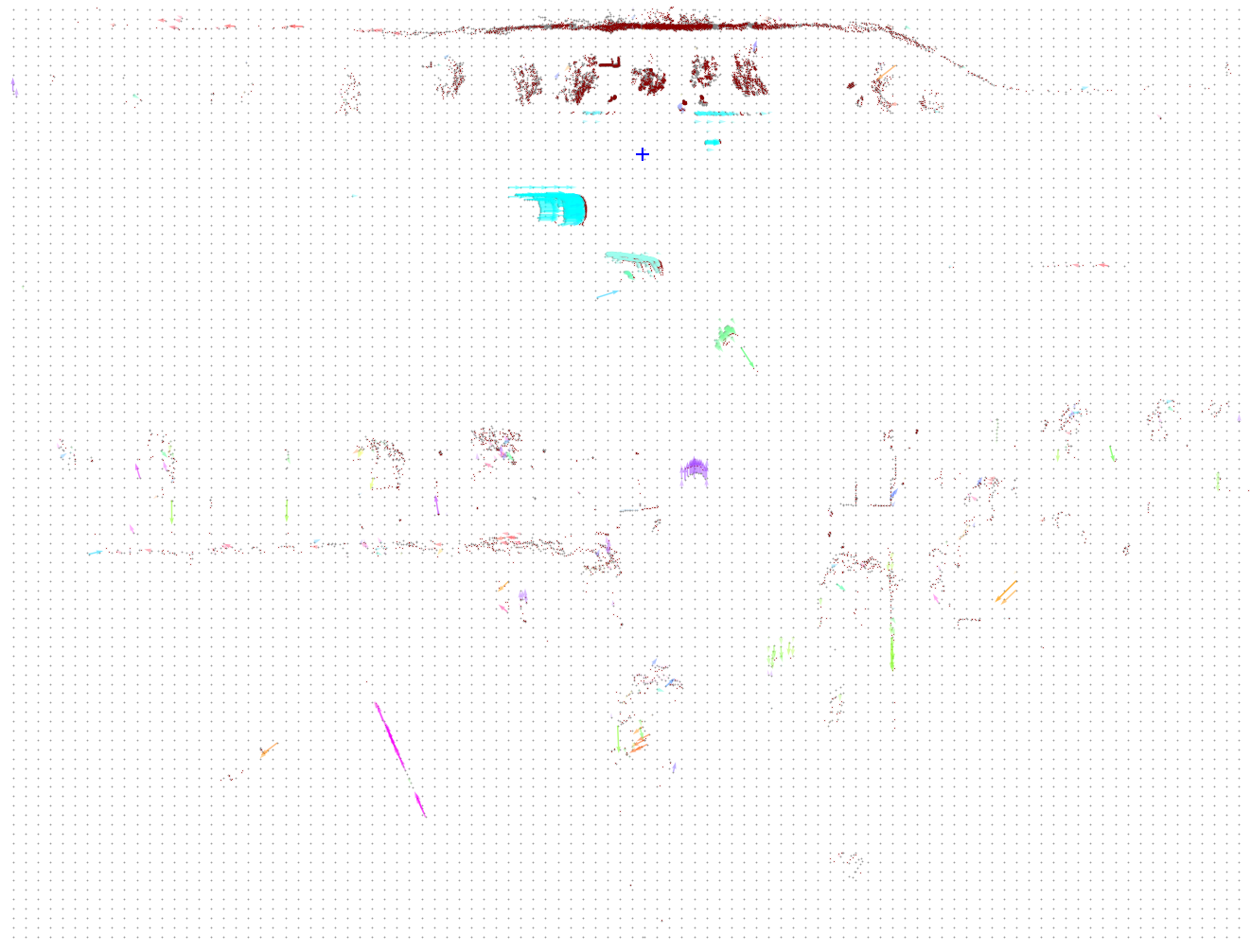}
    \end{subfigure}
    \begin{subfigure}[b]{0.42\textwidth}    
        \centering
        \includegraphics[width=0.9\textwidth]{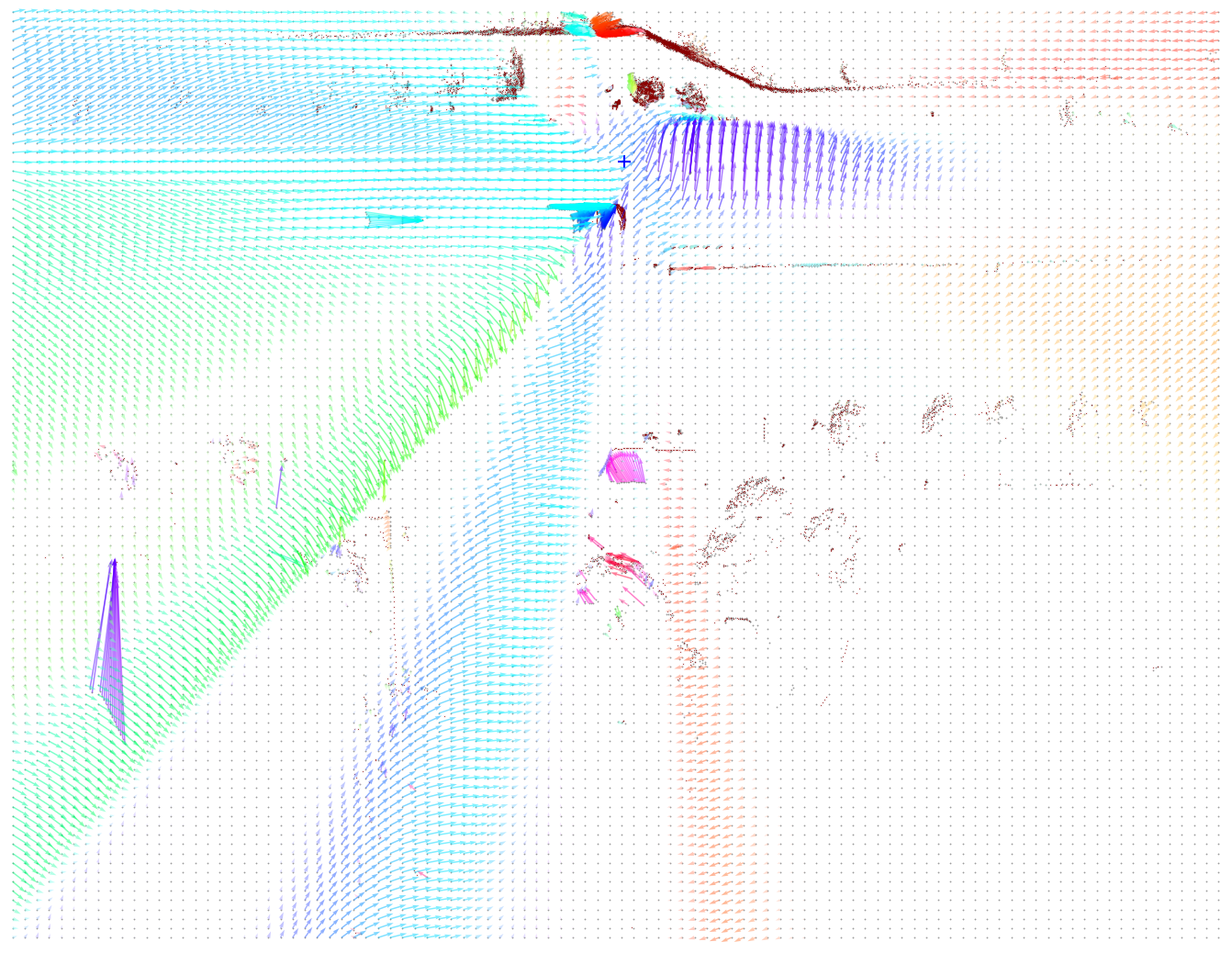}
    \end{subfigure}    
    \begin{subfigure}[b]{0.42\textwidth}    
        \includegraphics[width=0.9\textwidth]{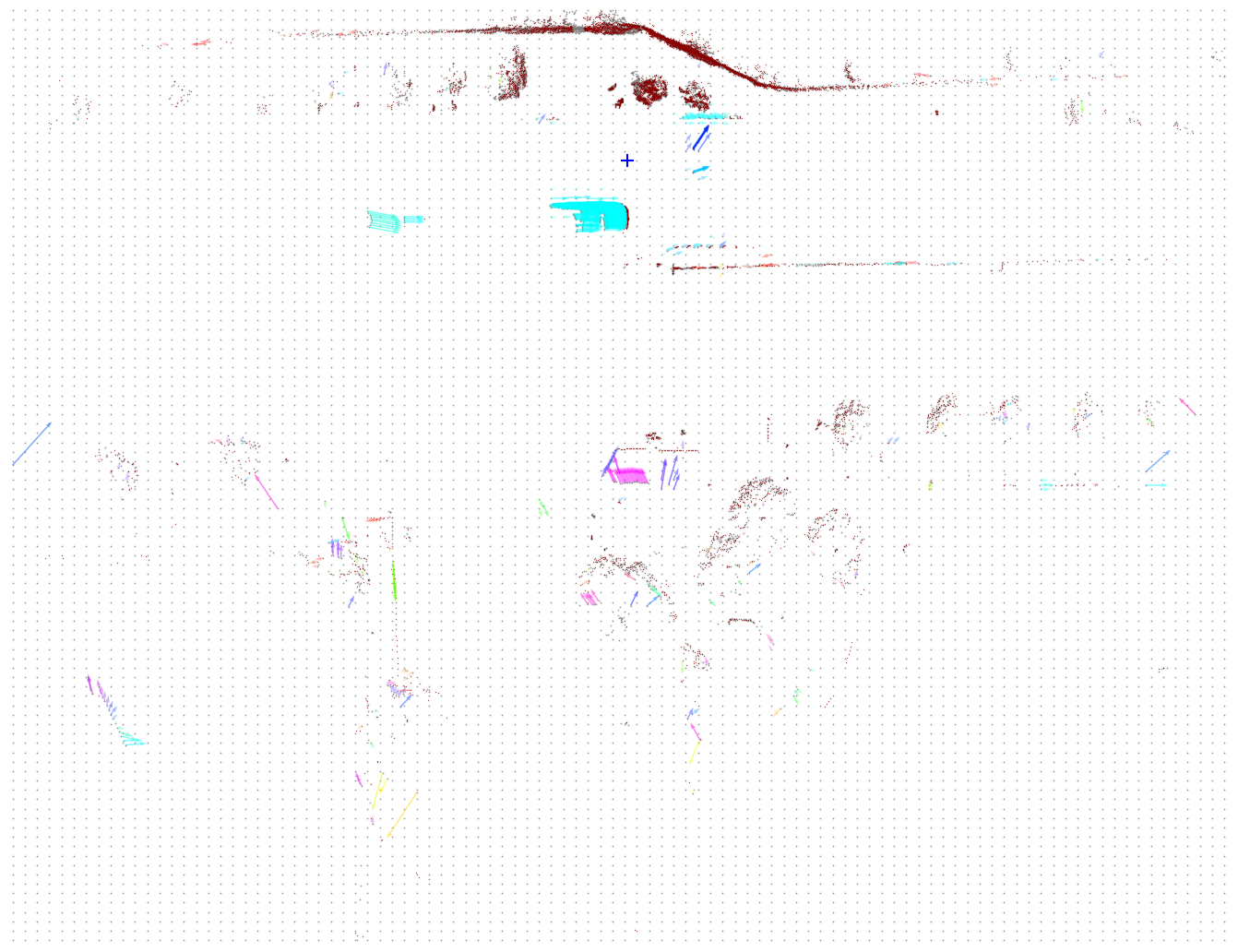}
    \end{subfigure}
    \begin{subfigure}[b]{0.42\textwidth}    
        \centering
        \includegraphics[width=0.9\textwidth]{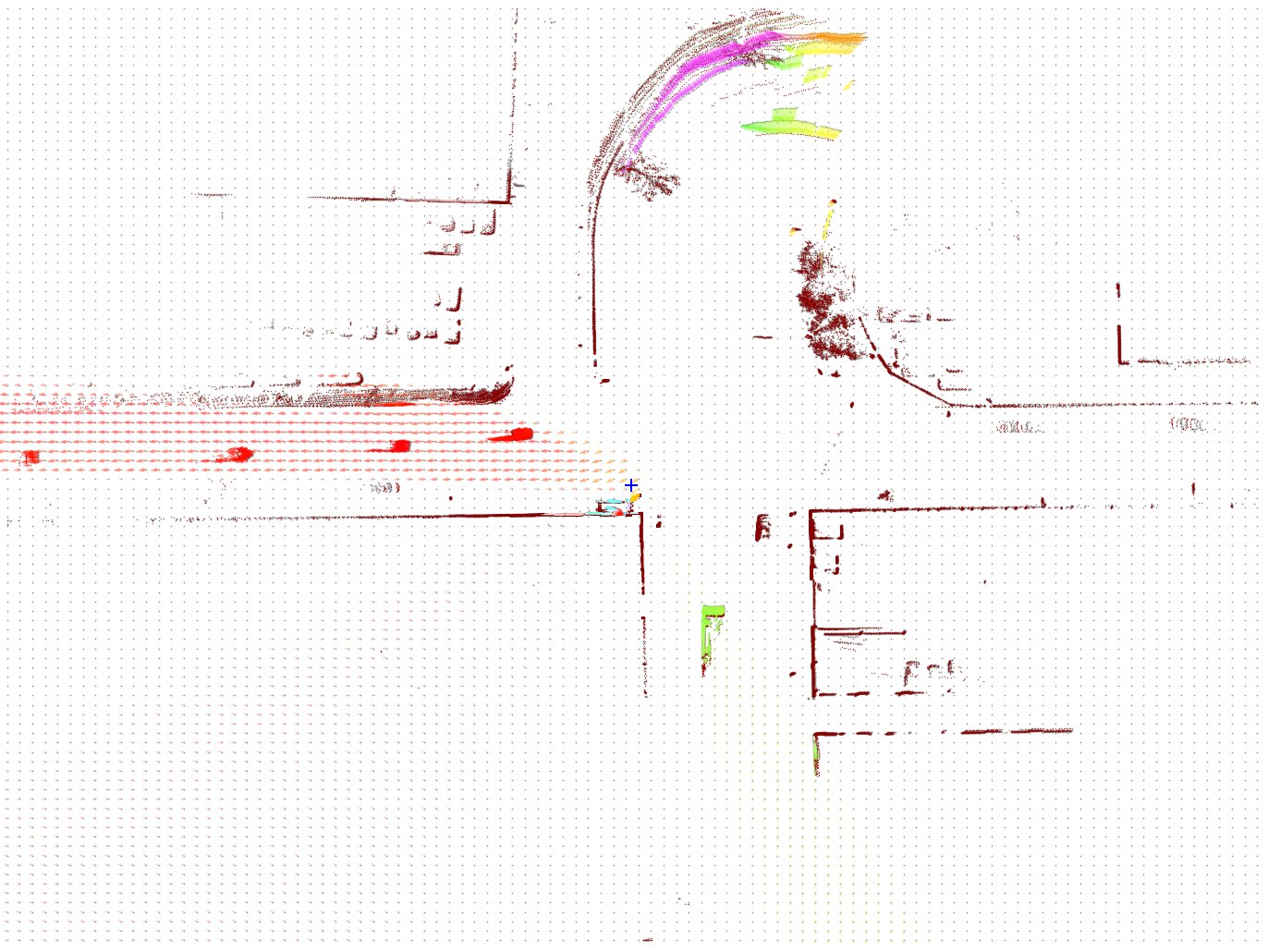}
        \caption{FNSF}
    \end{subfigure}
    \begin{subfigure}[b]{0.42\textwidth}    
        \includegraphics[width=0.9\textwidth]{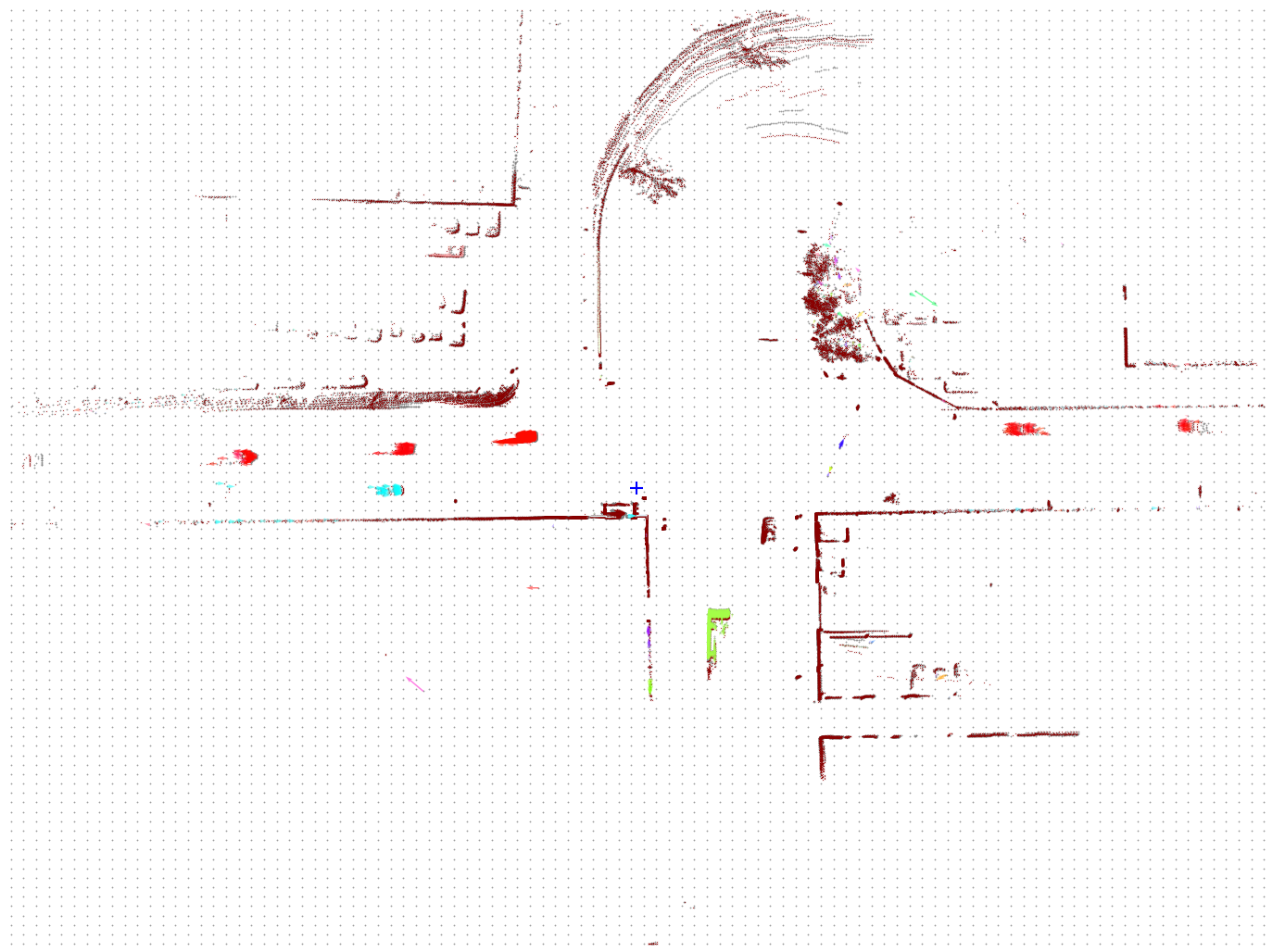}
        \caption{\floxels}
    \end{subfigure}
    \caption{\textbf{Comparison of scene flow fields after convergence}. We show a birds-eye view of the scene flow fields for FNSF (left) and Floxels (right) after convergence. Points at time $t$ are black, and $t+1$ are red. Floxels does well at isolating the dynamic environment whereas FNSF struggles to do so. Consequently, FNSF sometimes predicts zero-flow on dynamic objects and noisy flow vectors in the static regions as depicted above.
    }
    \label{fig:flow_opt_vis_converged}
\end{figure*}

We show further examples with a qualitative comparison of FNSF and Floxels in Fig. \ref{fig:flow_opt_vis_converged}.

\subsection{A closer look at the neural prior}
\begin{figure*}[t]
    \centering
    \begin{subfigure}[b]{1.0\textwidth}
        \centering
        \includegraphics[width=0.9\textwidth]{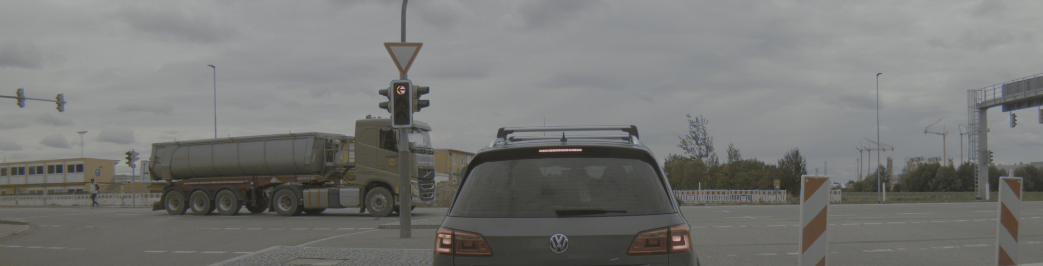}
        \caption{Matching camera image to scene flow fields}
    \end{subfigure}
    \begin{subfigure}[b]{1.0\textwidth}
        \centering
        \includegraphics[width=0.24\textwidth]{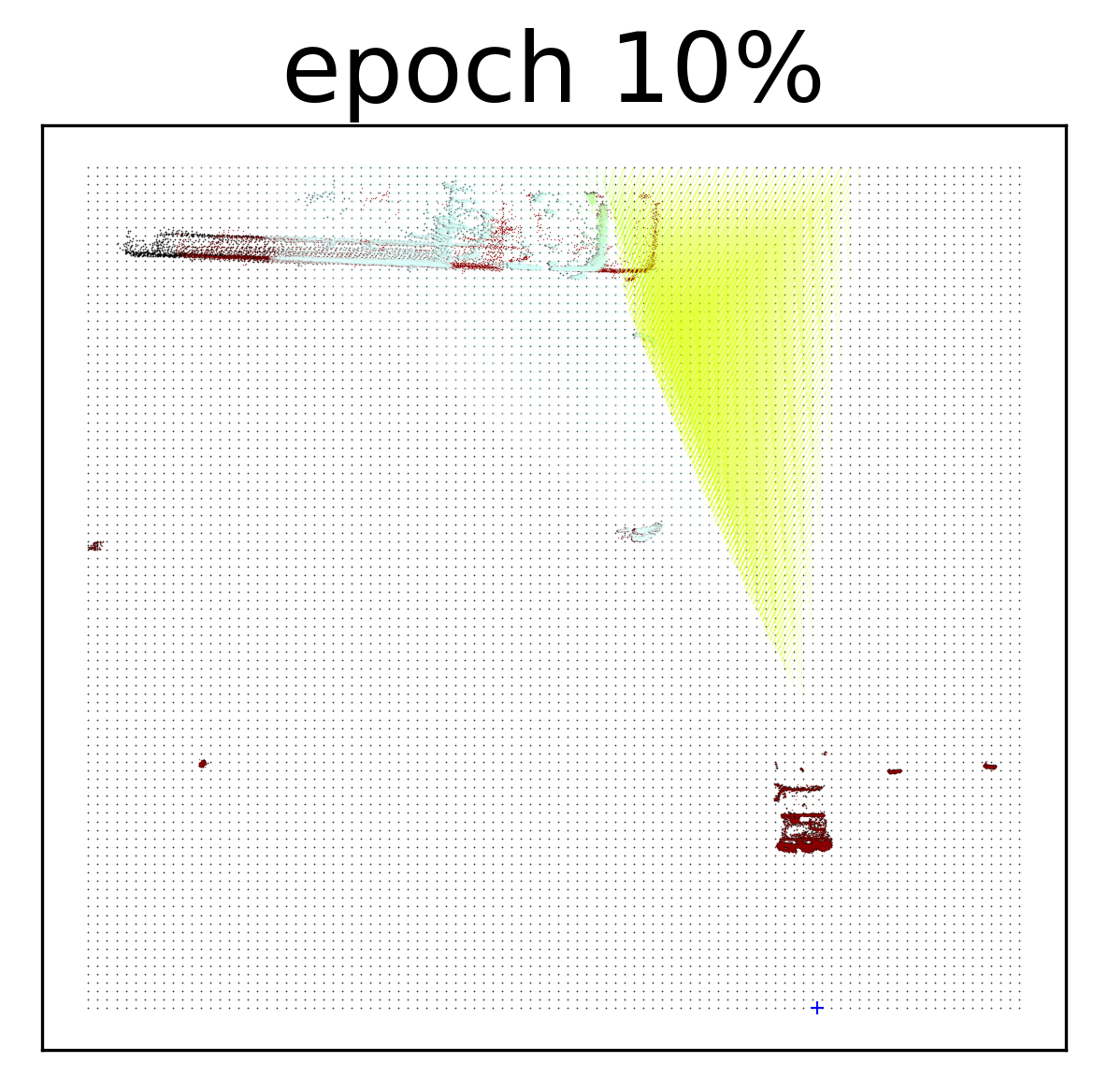}
        \includegraphics[width=0.24\textwidth]{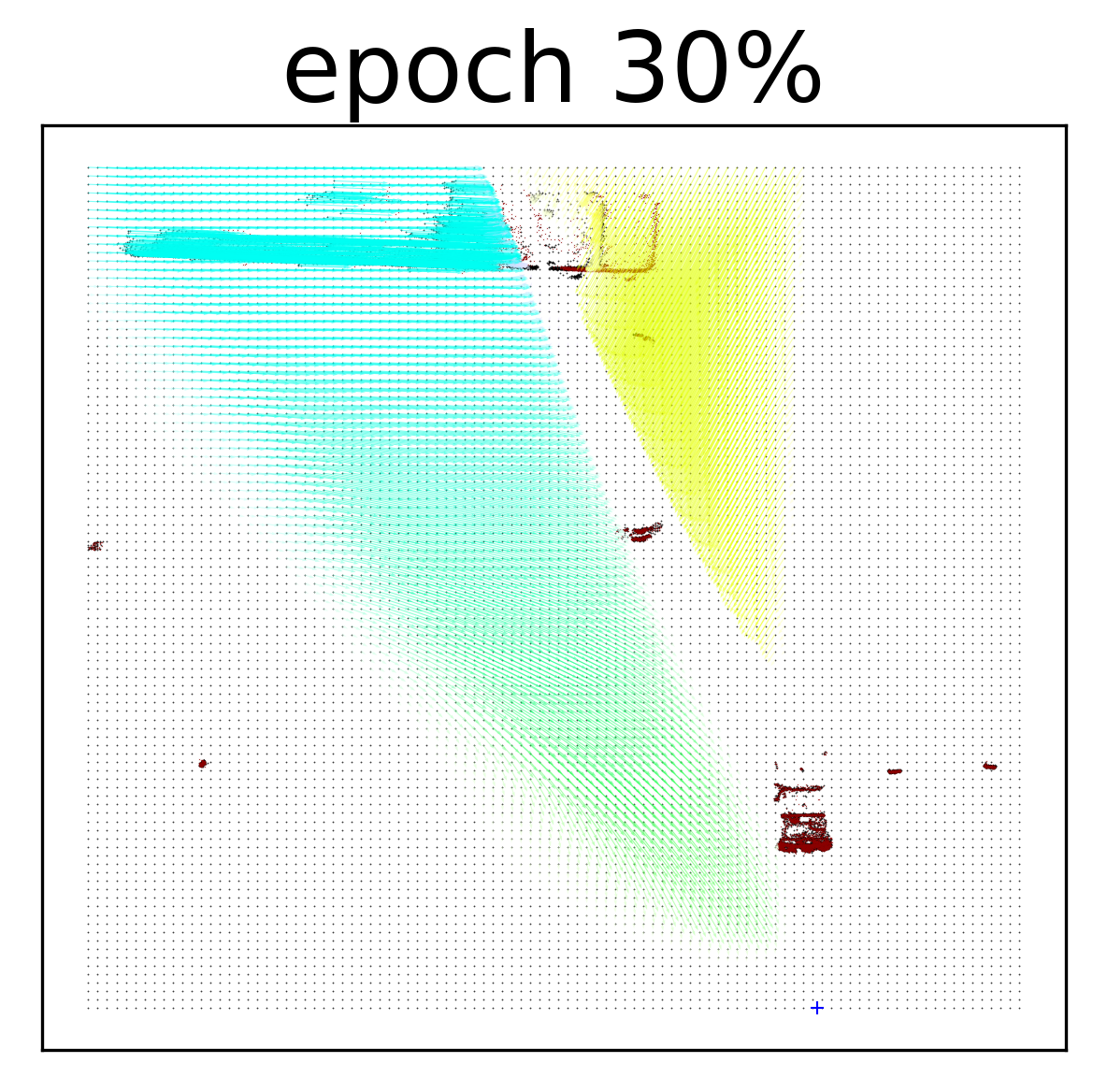}
        \includegraphics[width=0.24\textwidth]{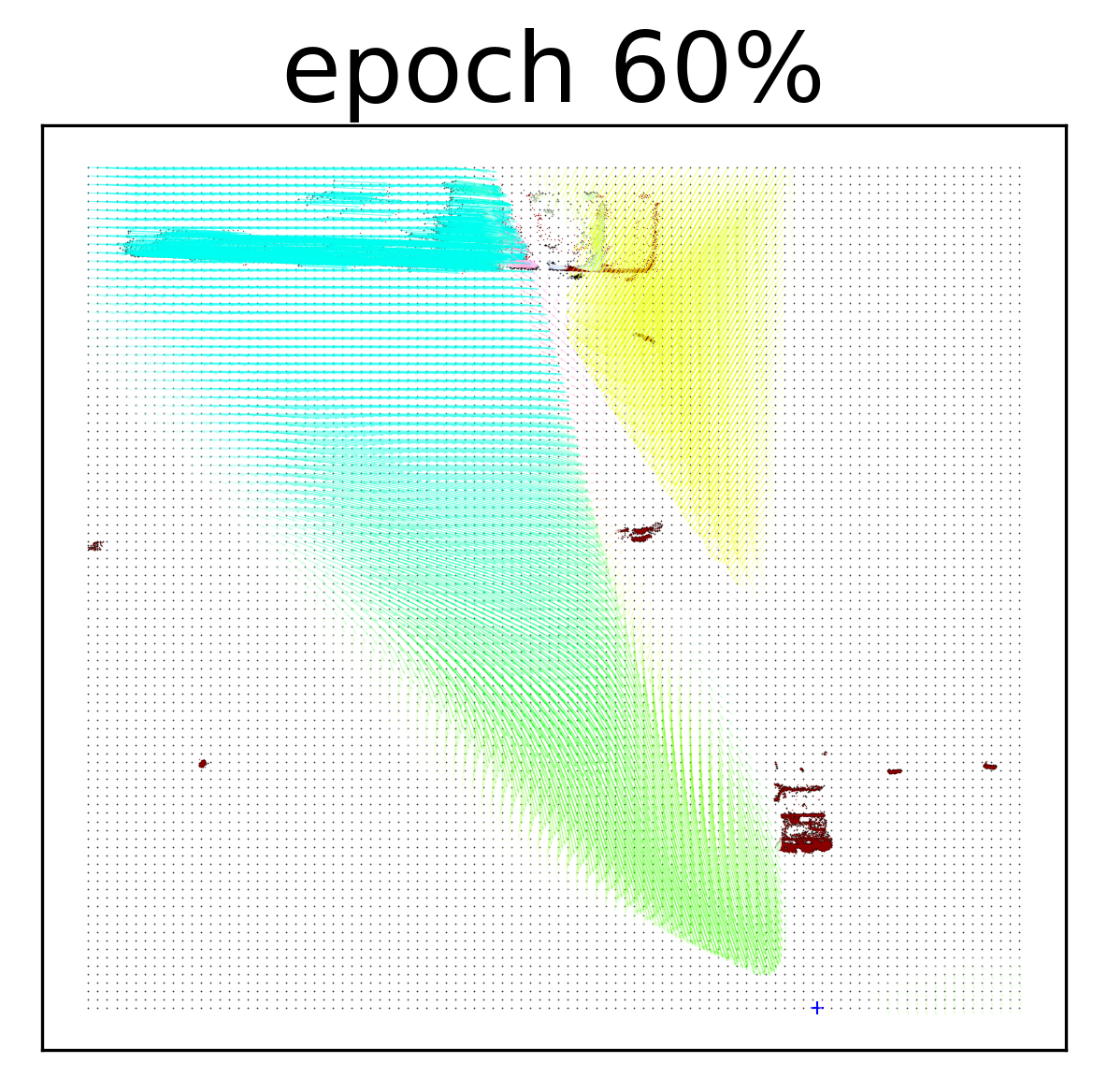}
        \includegraphics[width=0.24\textwidth]{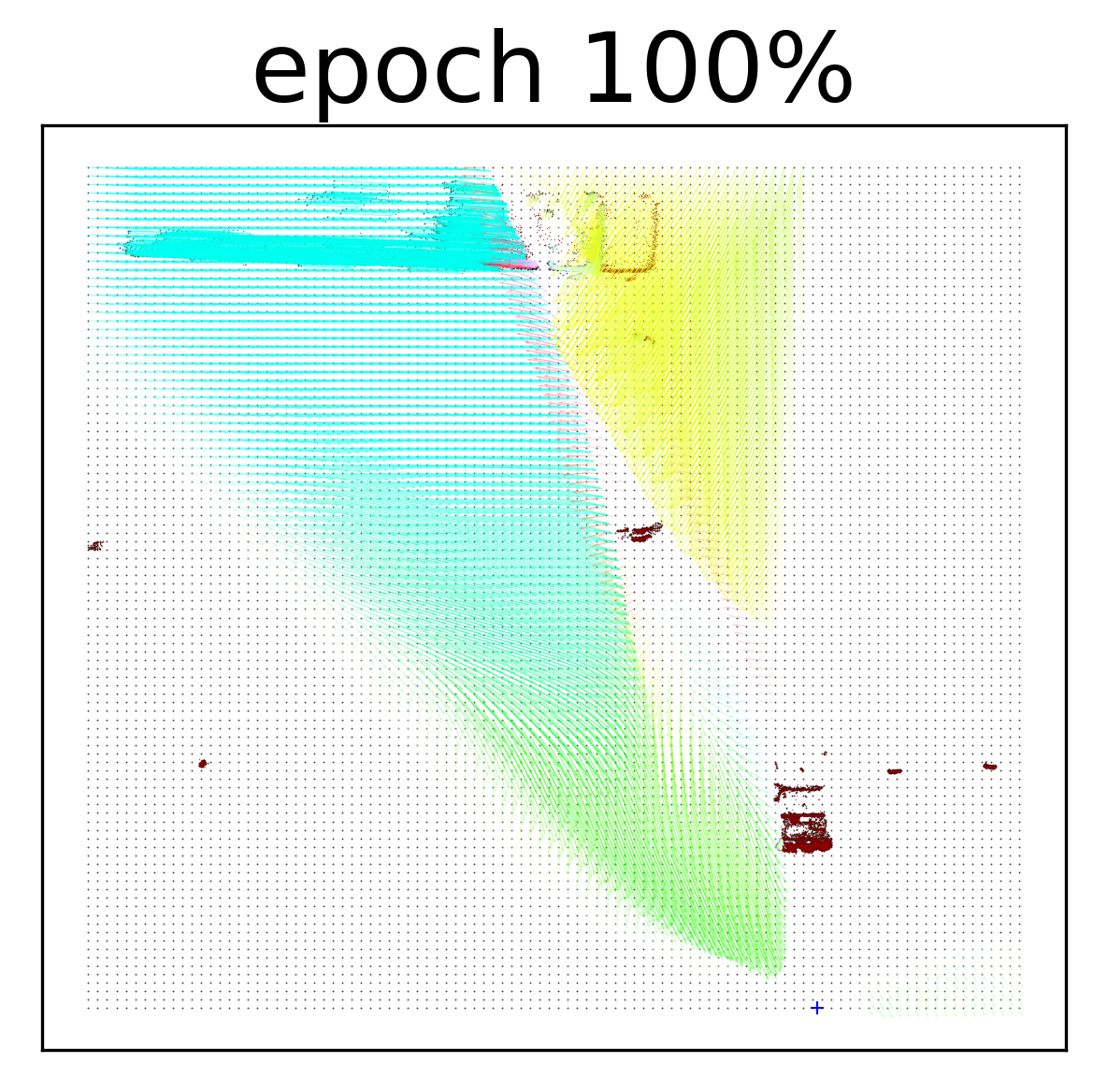}
        \caption{Original FNSF}
    \end{subfigure}
    \begin{subfigure}[b]{1.0\textwidth}
        \centering
        \includegraphics[width=0.24\textwidth]{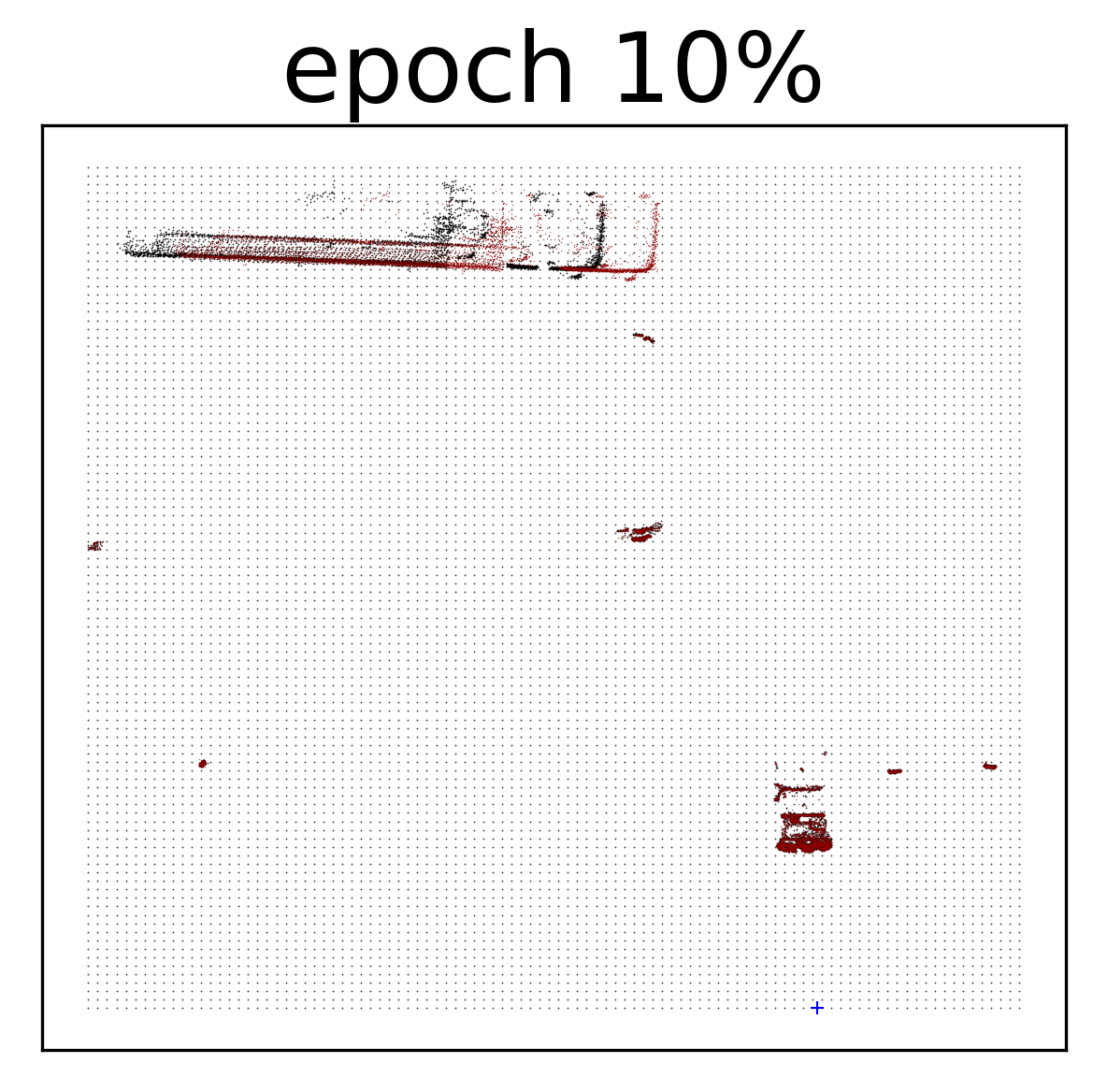}
        \includegraphics[width=0.24\textwidth]{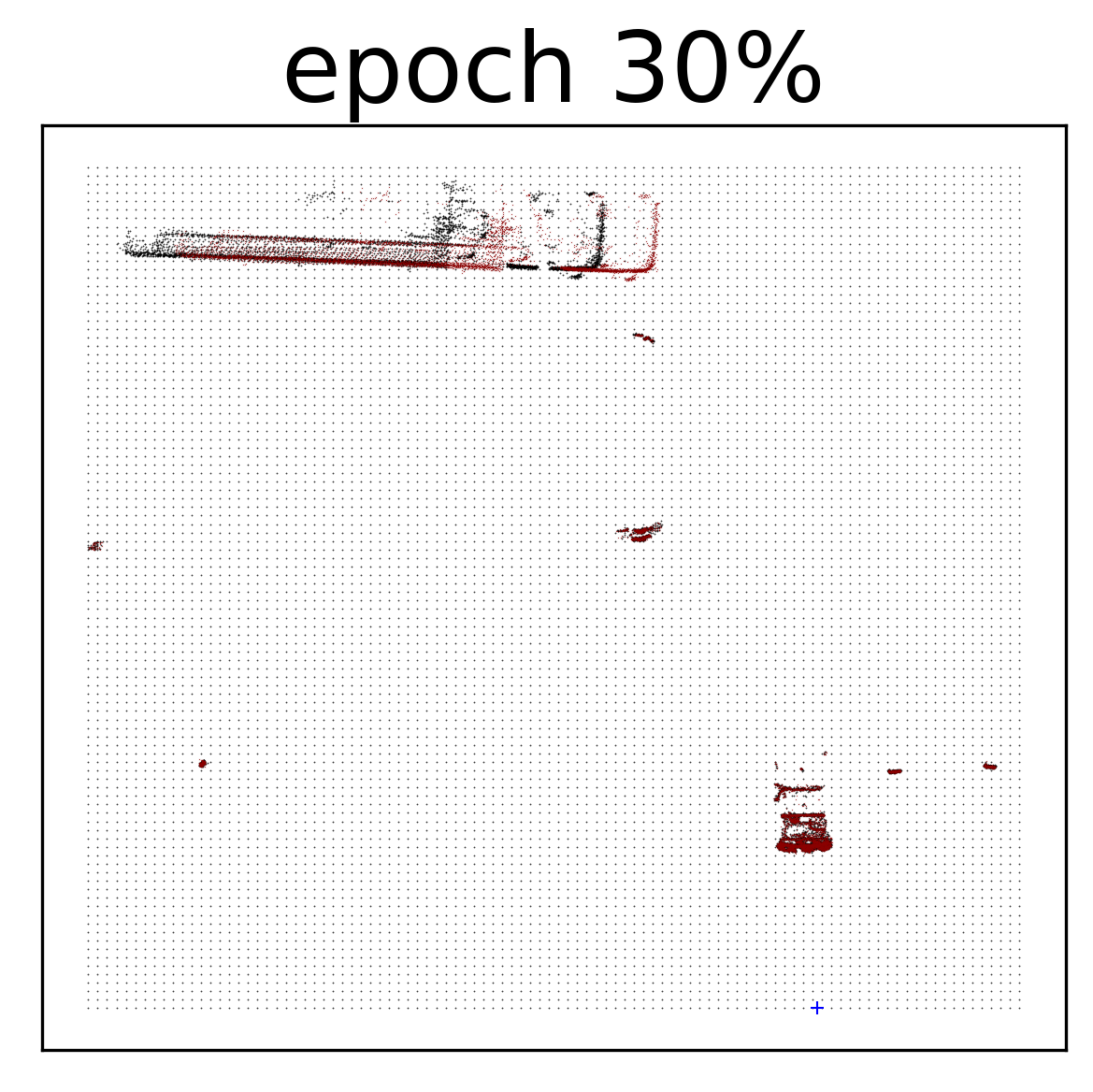}
        \includegraphics[width=0.24\textwidth]{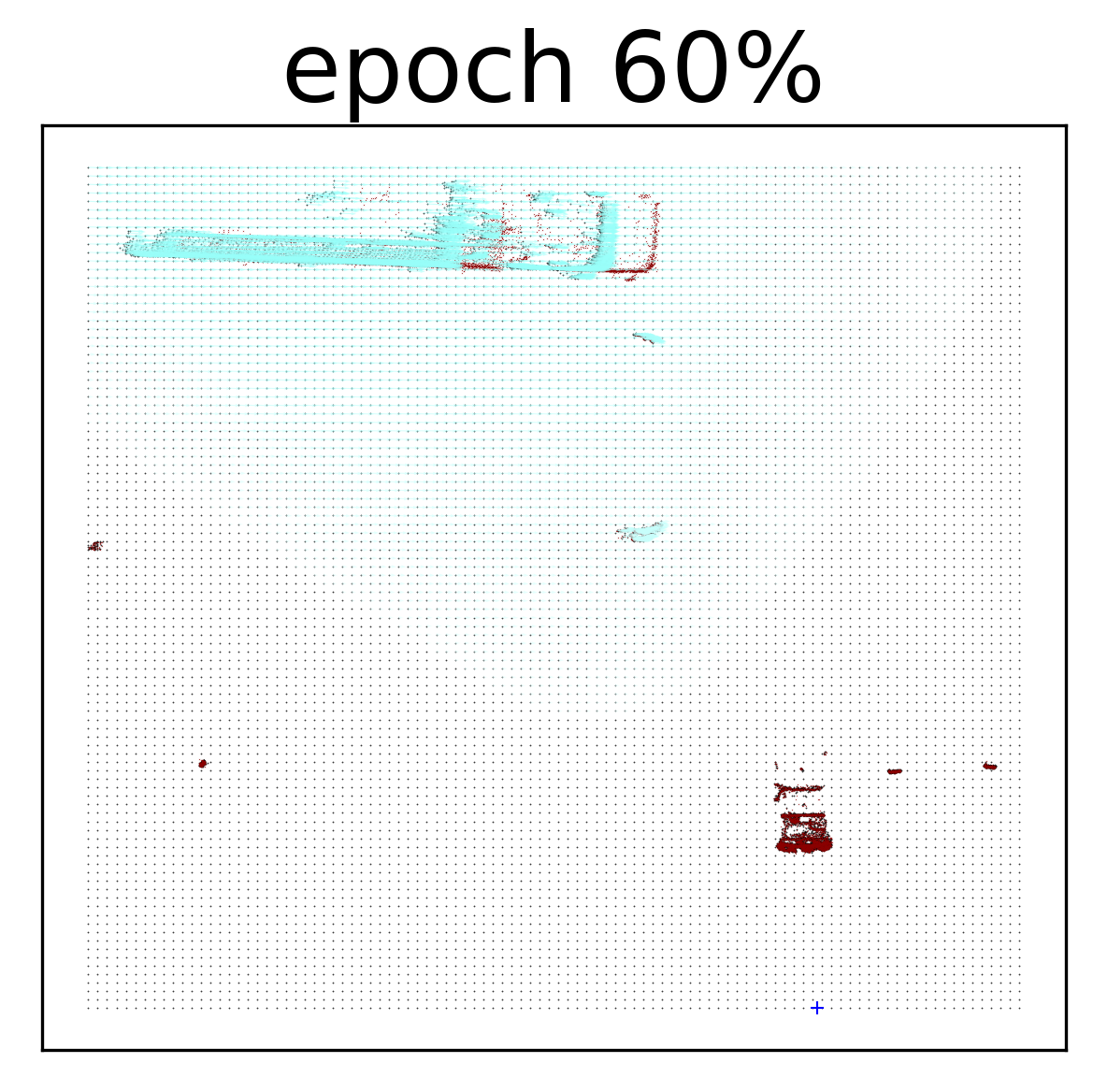}
        \includegraphics[width=0.24\textwidth]{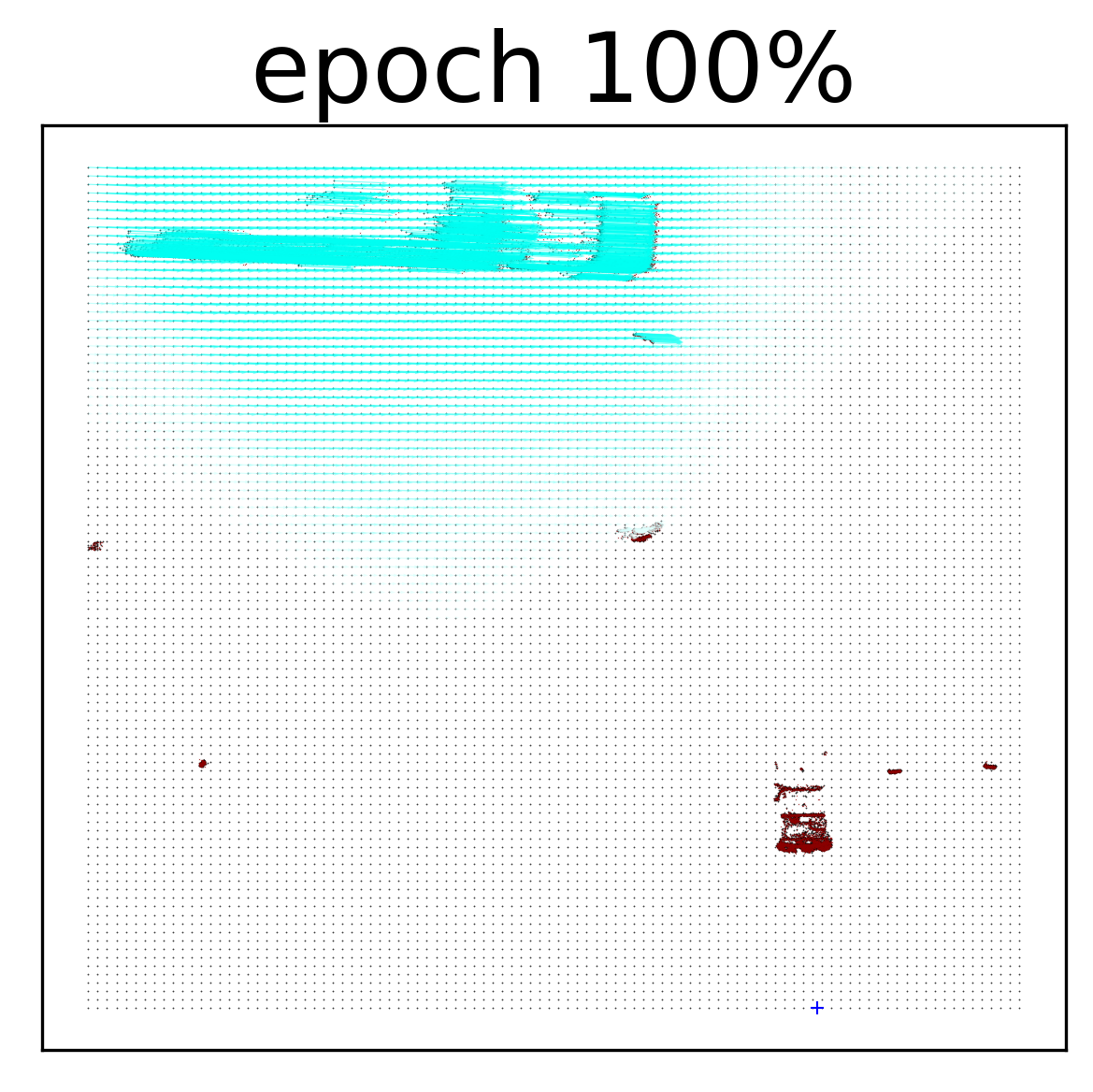}
        \caption{FNSF MLP with Floxels losses}
        \label{sub_fig:flow_opt_vis_fnsf_w_MLP}
    \end{subfigure}
    \begin{subfigure}[b]{1.0\textwidth}
        \centering
        \includegraphics[width=0.24\textwidth]{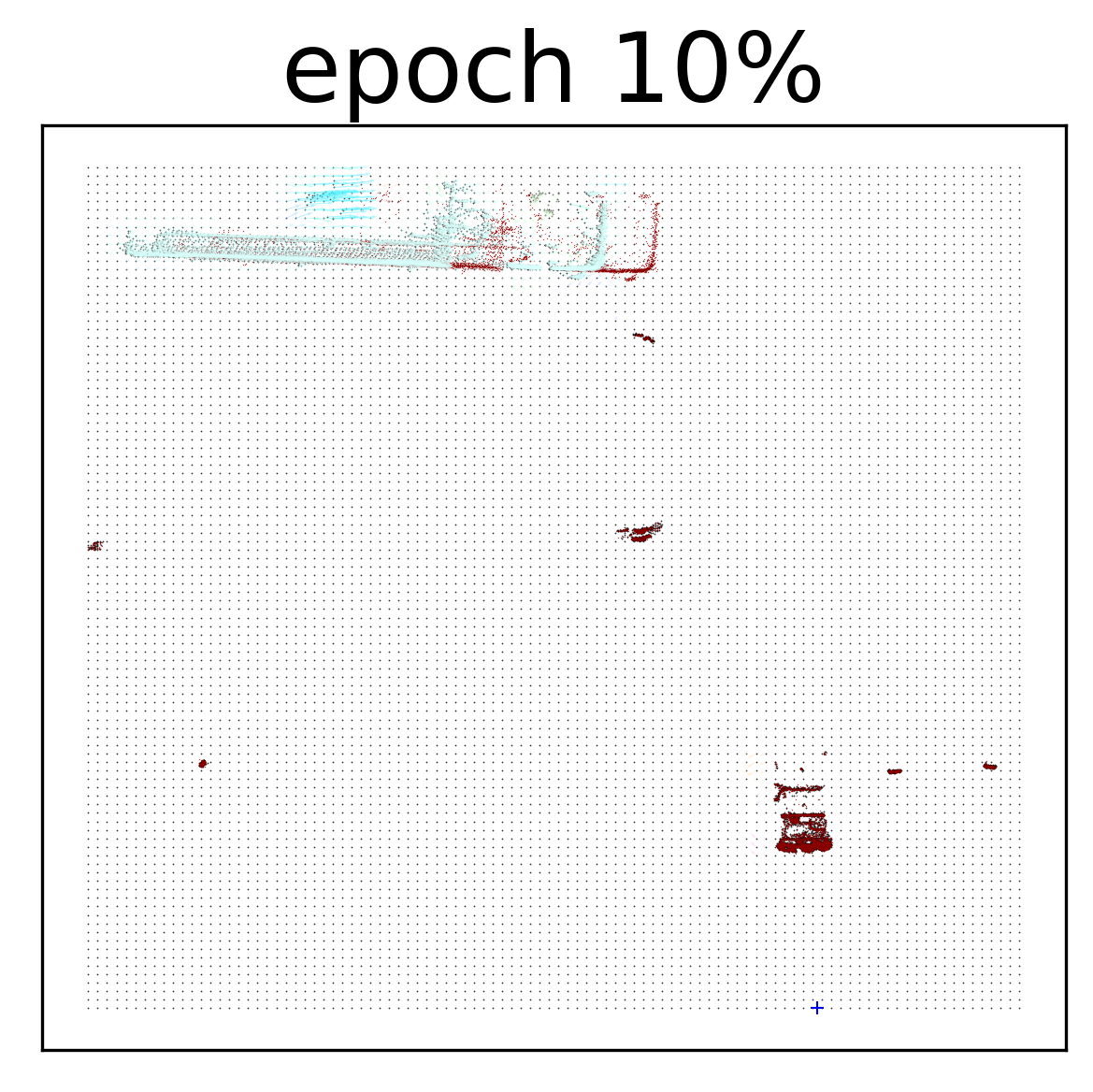}
        \includegraphics[width=0.24\textwidth]{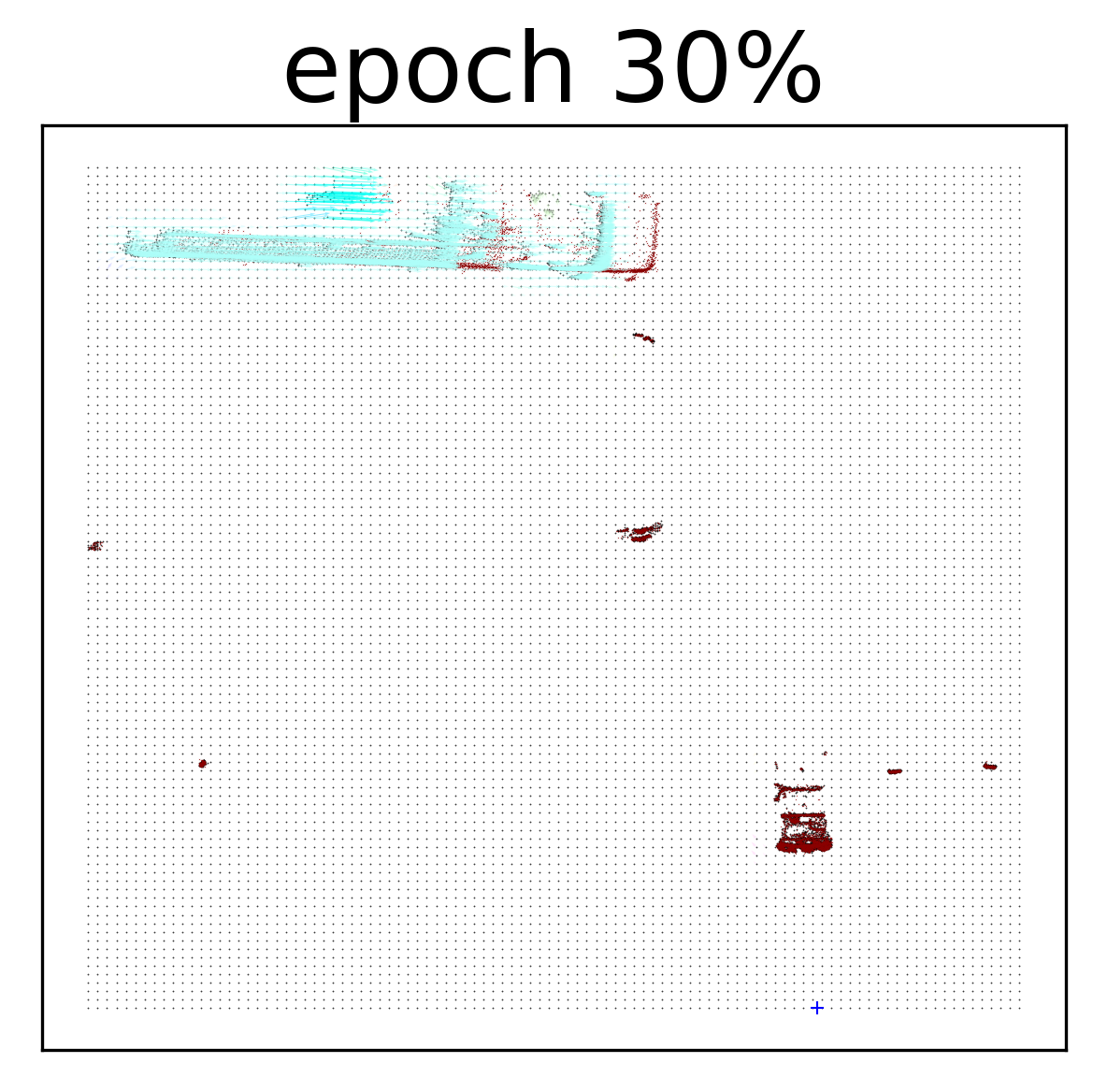}
        \includegraphics[width=0.24\textwidth]{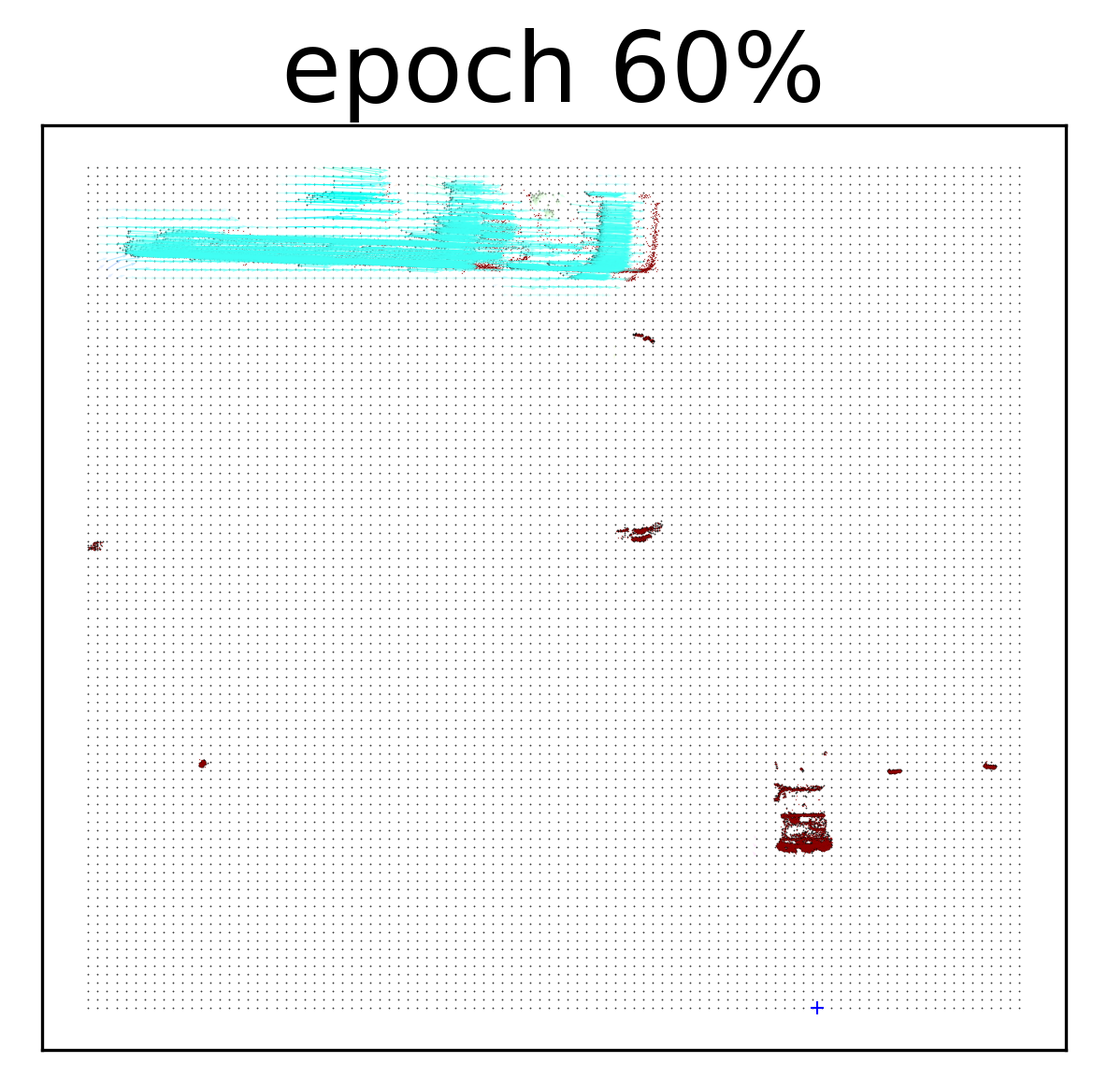}
        \includegraphics[width=0.24\textwidth]{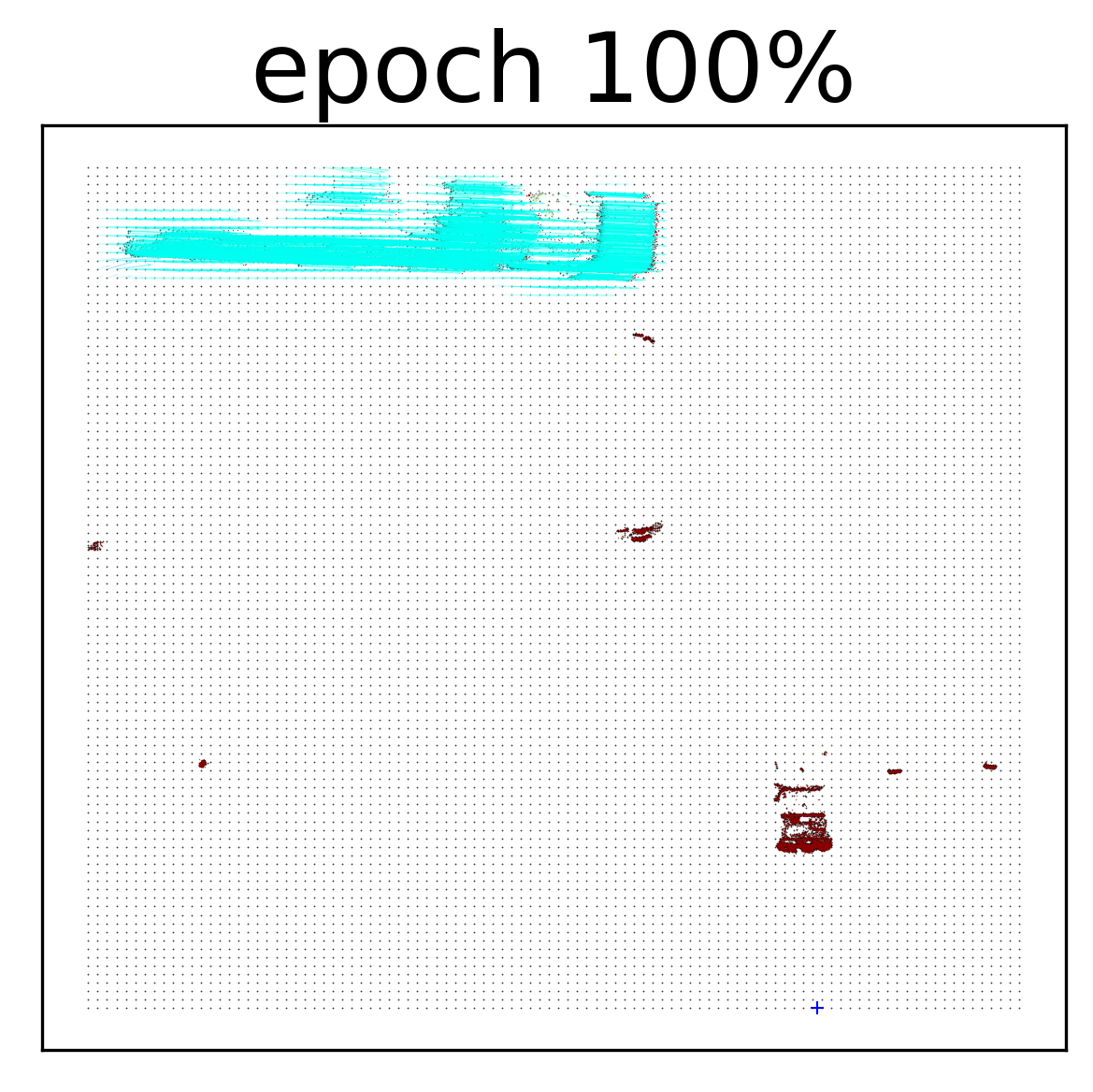}
        \caption{Floxels}
    \end{subfigure}
    \caption{\textbf{Evolution of scene flow comparison between FNSF, FNSF with \floxels losses and \floxels}. We show a birds-eye view of the estimated flow during optimization. FNSF exhibits problems in occluded regions and strong ``windmill artifacts''. For FNSF MLP with \floxels losses we observe that multi-frame and cluster losses help in occluded regions. Full \floxels also predicts zero-flow in empty regions and converges faster. Points at time $t$ are black and $t+1$ are red. Other colors are scene flow.
    }

    \label{fig:flow_opt_vis_fnsf}
\end{figure*}

In Sec.\ref{sec:neural_prior_good}, we compare qualitatively the influence of an MLP-based method and \floxels. To further separate the effects of the MLP vs. the voxel grid, we train an MLP using the \floxels losses. 
Fig.~\ref{fig:flow_opt_vis_fnsf} reveals that both slower convergence and windmill artifacts are consequences of the MLP and are primarily solved by the voxel grid. 
Nevertheless, it can be seen in \cref{sub_fig:flow_opt_vis_fnsf_w_MLP} that windmill artifacts are less pronounced when using the \floxels losses to train the MLP in comparison to the FNSF losses.
Furthermore, equipped with the multi-frame \floxels loss, the MLP can predict the flow in the challenging occluded region.
Together, these results highlight that both the voxel grid and the \floxels losses contribute to the superior performance of \floxels and solve different failure cases.

\subsection{Convergence speed and optimization videos}
Also visually, the convergence speed is much faster and more stable. We provide various videos of the optimization progress here: \url{https://www.youtube.com/playlist?list=PLCtNe14NZWtVjaoW_KDc19Kb-oThHNA2S}. We would like to explicitly highlight the differences between ``Flow Field evolution for Floxels'' and ``Flow Field Evolution FNSF''. Further, we would like to highlight the difficulties in removing the windmill artifacts, from MLP + \floxels losses (``Flow Field Evolution Custom MLP with Floxel Losses'').

\end{document}